\documentclass[final,5p,times,twocolumn,authoryear]{elsarticle} 

\usepackage{amsmath,amssymb}
\usepackage{graphicx}
\usepackage{booktabs}
\usepackage{multirow}
\usepackage{makecell}
\usepackage{tabularx}
\usepackage{siunitx}
\usepackage[table]{xcolor}
\usepackage{microtype}
\usepackage[hyphens]{url}
\usepackage{verbatim}
\usepackage[hidelinks]{hyperref}
\usepackage[capitalize,nameinlink]{cleveref} 
\usepackage{pifont}
\usepackage{float}
\usepackage[utf8]{inputenc}
\usepackage{amsmath,amssymb}
\usepackage[T1]{fontenc}
\usepackage{textcomp}
\usepackage{newtxtext,newtxmath}
\usepackage{amsmath,amssymb}
\usepackage[hyphens]{url}
\usepackage{subcaption}
\usepackage{pifont} 
\DeclareUnicodeCharacter{2212}{-}         
\DeclareUnicodeCharacter{00B0}{$^\circ$}  
\DeclareUnicodeCharacter{00D7}{$\times$}  
\DeclareUnicodeCharacter{2192}{$\to$}     
\DeclareUnicodeCharacter{00B1}{$\pm$}     
\DeclareUnicodeCharacter{2013}{--}        
\DeclareUnicodeCharacter{2014}{---}       
\DeclareUnicodeCharacter{201C}{``}        
\DeclareUnicodeCharacter{201D}{''}        

\journal{Computers in Biology and Medicine}

\newcommand{\method}{\textsc{LUX}} 

\begin{document}

\begin{frontmatter}

\title{LUX: A Lesion-Aware Graph-Conditioned Visual--Language Architecture for Explainable Endoscopic Captioning}

\author[aff1]{Alexis Iv\'an L\'opez Escamilla}

\author[aff1]{Gilberto Ochoa-Ruiz}\ead{gilberto.ochoa@tec.mx}\corref{cor1}
\author[aff1]{Salvador  Hinojosa}
\author[aff2]{Sharib Ali}

\affiliation[aff1]{organization={School of Engineering and Sciences, Tecnologico de Monterrey},
  city={Monterrey},
  postcode={64700},
  state={NL},
  country={Mexico}}

\affiliation[aff2]{organization={  AI in Medicine and Surgery group, School of Computer Science, University of Leeds},
  city={Leeds},
  postcode={LS2 9JT},
  country={United Kingdom}}

\cortext[cor1]{Corresponding author.}

\begin{abstract}
The interpretation of endoscopic imagery in ulcerative colitis remains a complex and subjective process, often affected by variability in human assessment and the subtle presentation of mucosal inflammation. Although deep learning has enabled remarkable advances in automated analysis, most current vision--language models describe endoscopic scenes through global embeddings that neglect the localized and relational nature of pathological evidence. This limitation constrains both clinical reliability and interpretability, as the linguistic descriptions generated by such models are rarely grounded in explicit visual reasoning. 

To address this challenge, we introduce LUX (Lesion-aware Unified eXplainable captioning), a lesion-aware graph-conditioned visual--language architecture designed to bridge visual understanding and linguistic interpretation in endoscopic image captioning. LUX constructs a lesion-centric scene graph from Grad-CAM and CBAM activation maps, representing pathological regions as structured nodes and encoding their spatial and clinical relations within a graph topology. These graph embeddings are subsequently integrated into the cross-attention layers of a T5 decoder, allowing each word in the generated caption to attend to specific lesion nodes rather than to abstract global features. Through this design, the model establishes a direct alignment between linguistic semantics and pathological evidence, enabling token-level interpretability and relational reasoning. 

Experimental results demonstrate that LUX consistently outperforms strong baseline and state-of-the-art medical captioning models across BLEU, METEOR, ROUGE-L, and CIDEr metrics, with particularly strong gains in CIDEr, indicating improved semantic alignment with expert-authored descriptions. In addition, LUX significantly reduces hallucinated clinical findings and improves lesion-level grounding, as evidenced by more faithful correspondence between generated tokens and localized pathological regions.

By unifying lesion localization, graph-based representation, and contextual language generation within a single architecture, LUX offers a transparent and clinically meaningful step toward explainable artificial intelligence for gastrointestinal endoscopy.
\end{abstract}


%

\end{frontmatter}

%
%
\section{Introduction}

\textbf{Ulcerative colitis (UC)} is a chronic inflammatory bowel disease characterized by recurrent mucosal inflammation that varies in severity and spatial distribution throughout the colon \citep{ref_uc_pathology}. Endoscopic evaluation remains the clinical standard for assessing disease activity and therapeutic response \citep{ref_endoscopy_diagnosis}. However, this process is inherently subjective, requiring clinicians to interpret subtle vascular, textural, and color variations under heterogeneous imaging conditions. Even with standardized grading systems such as the \textbf{Mayo Endoscopic Subscore (MES)}, substantial inter-observer variability persists, particularly in borderline or moderate cases \citep{ref_mes_variability}. Consequently, considerable effort has been devoted to developing computational methods capable of improving diagnostic reproducibility and supporting objective endoscopic assessment.

Recent advances in \textbf{computer vision} and \textbf{deep learning} have accelerated the development of computer-aided diagnosis systems for gastrointestinal endoscopy \citep{ref_cv_dl_endoscopy}. Convolutional neural networks and transformer-based architectures have demonstrated strong performance in lesion detection, segmentation, and severity classification \citep{ref_transformers_medical}. Nevertheless, most existing approaches remain focused on classification-oriented objectives and provide limited insight into the semantic or pathological reasoning underlying model predictions. Their outputs are commonly restricted to categorical scores or confidence values, which constrains interpretability and limits clinical trust.

\begin{figure}[t]
    \centering
    \includegraphics[width=0.45\textwidth]{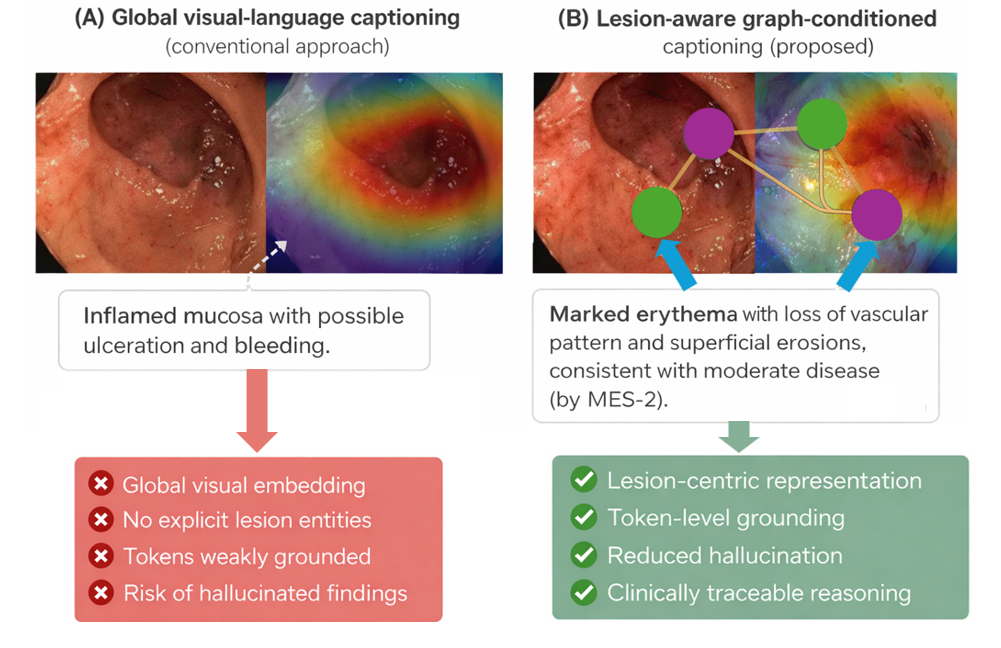}
    \caption{Comparison between conventional global visual-language captioning and the proposed lesion-aware graph-conditioned approach.
    \textbf{(A)} Conventional medical captioning compresses visual information into global embeddings, often producing weakly grounded descriptions and hallucinated findings.
    \textbf{(B)} In \method{}, pathological regions are represented as lesion nodes within a relational graph that conditions decoder cross-attention, enabling token-level grounding and clinically traceable reasoning.}
    \label{fig:comparison}
\end{figure}

The emergence of \textbf{vision--language models (VLMs)} has introduced new possibilities for generating natural-language descriptions directly from medical imagery \citep{ref_vlm_medical}. In clinical settings, image captioning offers the potential to bridge visual findings and diagnostic reasoning by producing structured textual descriptions aligned with expert interpretation. However, most existing medical captioning approaches rely on global visual embeddings that compress heterogeneous pathological patterns into a single representation. As a result, generated captions frequently lack lesion-level grounding, conflate distinct findings, or introduce unsupported pathological statements \citep{ref_captioning_limitations}. These limitations are illustrated in Fig.~\ref{fig:comparison}.

At the same time, the growing adoption of \textbf{large language models (LLMs)} in medical vision--language research has intensified concerns regarding interpretability, scalability, and clinical grounding \citep{ref_llm_healthcare}. Although architectures such as T5 and multimodal transformers provide remarkable linguistic fluency, their outputs often depend heavily on large-scale language priors rather than explicit pathological evidence. In medical imaging domains where annotated data remain limited and interpretability is essential, there is a critical need for architectures capable of maintaining clinical transparency without relying exclusively on massive language models or weakly grounded global representations.

These challenges are closely related to the broader problem of \textbf{explainable artificial intelligence (XAI)} in healthcare \citep{ref_xai_biomedical}. Existing explainability approaches, including Grad-CAM and attention visualization, provide post hoc interpretations that highlight influential image regions after prediction. While useful, such techniques do not explicitly integrate interpretability into the reasoning process itself. In clinical practice, physicians require systems that can associate localized pathological findings with coherent diagnostic descriptions rather than merely highlighting diffuse attention regions. This distinction is particularly important in ulcerative colitis endoscopy, where disease severity depends on the spatial co-occurrence and progression of multiple mucosal abnormalities.

\begin{figure*}[t]
    \centering
    \includegraphics[width=0.9\textwidth]{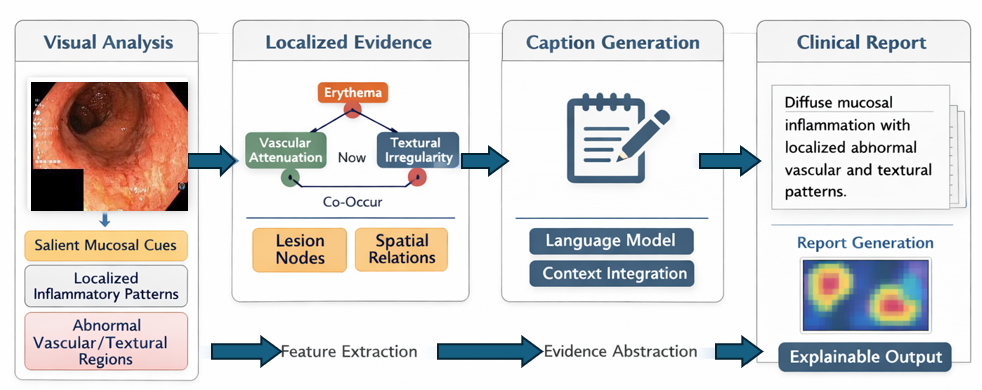}
    \caption{Illustration of the evidence-first reasoning pipeline in the proposed \method{} framework. Raw endoscopic imagery is progressively transformed into lesion-level representations and structured through a scene--lesion graph that explicitly conditions language generation. This design emphasizes grounded multimodal reasoning rather than direct image-to-text prediction.}
    \label{fig:lux_conceptual}
\end{figure*}

Prior lesion-aware frameworks \citep{lopezescamilla2025lesionaware} partially addressed this limitation by incorporating localized visual grounding into ulcerative colitis assessment. Although these approaches improved feature-level interpretability, they did not explicitly model relations among lesion entities nor directly constrain language generation through relational reasoning. Consequently, the generated descriptions remained only partially connected to the underlying pathological evidence.

To address these limitations, we introduce \textbf{LUX (Lesion-aware Unified eXplainable captioning)}, a lesion-aware graph-conditioned visual--language architecture for explainable endoscopic image captioning. The proposed framework, illustrated in Fig.~\ref{fig:lux_conceptual}, models the endoscopic scene as a structured graph of pathological regions derived from Grad-CAM and CBAM activations. Lesion nodes encode localized pathological evidence, while graph edges capture spatial and semantic relationships among lesions. These graph representations are subsequently integrated into the cross-attention layers of a T5 decoder, enabling each generated token to attend jointly to visual embeddings and lesion-level relational structures.

Rather than treating caption generation as a direct image-to-text translation problem, \method{} formulates medical captioning as an evidence-grounded clinical reasoning process. The framework progressively localizes pathological findings, abstracts them into lesion entities, reasons over their relationships, and generates clinically structured language explicitly linked to visual evidence. Through this design, \method{} establishes traceable alignment between pathological regions and generated descriptions, reducing hallucinated findings while improving interpretability and semantic consistency.

The principal contributions of this work are summarized as follows:

\begin{itemize}
    \item We propose a lesion-aware graph-conditioned vision--language architecture that integrates localized pathological evidence directly into the caption generation process.
    
    \item We introduce a scene--lesion graph representation that models spatial and semantic relationships among inflammatory findings in ulcerative colitis endoscopy.
    
    \item We develop a dual cross-attention mechanism that enables token-level grounding between generated language and lesion-level representations.
    
    \item We demonstrate improved caption quality, lesion grounding, and hallucination reduction compared with conventional medical captioning baselines.
    
    \item We formulate endoscopic image captioning as an evidence-constrained multimodal reasoning problem aligned with clinical interpretation practices.
\end{itemize}

\subsection{Paper Organization}
\label{sec:paper_organization}

The remainder of this article is organized as follows. Section~2 reviews related work in medical image captioning, explainable AI, and graph-based reasoning. Section~3 presents the proposed methodology, including lesion-aware graph construction and graph-conditioned caption generation. Section~4 describes the datasets and preprocessing pipeline. Section~5 details the experimental setup and evaluation metrics, while Section~6 presents quantitative and qualitative results. Finally, Sections~7--9 discuss limitations, clinical implications, and conclusions.
\section{Related Work}
\label{sec:related}

\subsection{Medical Image Captioning: Clinical Domains and Methodological Advances}

Medical image captioning (MIC) has progressively evolved from early CNN--RNN architectures for image-to-text generation toward transformer-based and multimodal frameworks capable of producing more coherent and clinically structured descriptions \citep{jing2018automatic,boag2020radiology,li2020comparison,liu2021clinically,xue2022advancing,yuan2021automatic}. In radiology, where automated report generation has been most extensively studied, recent approaches have increasingly incorporated multimodal representations and structured clinical information to improve semantic consistency and reporting quality \citep{liu2023unified}.

Beyond radiology, medical captioning has expanded into pathology, ophthalmology, and dermatology through fine-grained, attention-guided, and domain-specific approaches such as PathCap \citep{wang2021pathcap}, HistoCap \citep{mitra2023histocap}, RetinaCap \citep{yan2022retina}, and clinically oriented skin and ophthalmic captioning systems \citep{he2020pathologycap,li2021skinreport,huang2023ophcap}. In parallel, multimodal pretraining frameworks such as MedicalCLIP and MedCLIP \citep{zhang2022medicalclip,yao2022medclip}, together with BLIP-based medical adaptations \citep{chen2023blipmed}, have demonstrated that improved alignment between visual and textual representations can enhance semantic consistency and transfer across medical imaging tasks. Retrieval-guided captioning, interpretable transformers, and hybrid alignment mechanisms have further explored ways to improve robustness and visual--linguistic correspondence \citep{yuan2023retrievalcap,wu2023medicaltransformer,meng2022hybridcap}.

Despite these advances, current MIC systems continue to face important limitations, including data scarcity, linguistic ambiguity, and a widespread dependence on global visual representations \citep{biswal2022review,srinivasan2023survey}. This limitation is particularly relevant in gastrointestinal endoscopy, which remains comparatively underexplored from a caption-generation perspective, with most existing computational approaches focusing primarily on classification or segmentation rather than grounded narrative generation \citep{min2021endoscopy,aslan2023colitisai}. Global image embeddings can capture overall appearance but provide limited correspondence between generated clinical statements and the localized pathological evidence supporting them.

The lesion-aware framework proposed by \citet{lopezescamilla2025lesionaware} partially addressed this gap by introducing localized visual grounding into ulcerative colitis captioning. However, the approach remained limited to feature-level alignment and did not explicitly represent relational interactions among lesion entities or directly constrain language generation through structured lesion-level reasoning. Consequently, generated descriptions remained only partially linked to the pathological evidence underlying the predicted clinical interpretation.

These limitations motivate the development of architectures that jointly integrate lesion localization, relational reasoning, and language generation. The proposed \method{} framework addresses this need by moving beyond global visual representations toward lesion-aware, graph-conditioned caption generation in which localized pathological findings and their relationships explicitly influence the generation process.

\subsection{Graph-Based and Relational Reasoning in Vision and Captioning}

Graph-based modeling has become increasingly relevant for representing structured relationships in visual reasoning tasks. Scene-graph approaches for image captioning \citep{yang2019autoencoding,li2022sggcaption} demonstrated that relational representations can improve semantic coherence by modeling interactions among visual entities. Similarly, relation-aware graph neural networks \citep{gu2021relationaware} enabled richer contextual reasoning across spatial regions.

Within medical imaging, graph neural networks (GNNs) have been applied to structured diagnostic inference, disease prediction, and multimodal reasoning \citep{kim2021medicalgnn,zhou2023graphmed,xie2022graphdiagnosis}. Additional surveys \citep{xu2023gnnreview,sun2023relational,yadav2023graphsurvey} highlighted the growing role of relational learning in biomedical AI, while causal relational attribution graphs \citep{chattopadhay2022crag} explored explainable graph-based visual reasoning.

Despite these advances, prior graph-based captioning approaches generally treat relational structures as encoder-side representations or auxiliary constraints. Decoder-level graph conditioning for lesion-aware medical caption generation remains comparatively unexplored, particularly in gastrointestinal endoscopy.

\subsection{Clinical Narrative Generation and Foundation Vision--Language Models}

Medical image analysis has progressively evolved from short descriptive captions toward the generation of structured clinical narratives. Early radiology report-generation systems demonstrated the feasibility of translating imaging findings into coherent diagnostic descriptions \citep{jing2018automatic,li2020comparison}, while subsequent transformer-based approaches improved linguistic fluency, contextual modeling, and factual consistency     \citep{chen2023blip}. These developments established automated report generation as an important extension of medical image captioning, with the potential to transform visual findings into clinically interpretable narratives. However, most of these advances have been concentrated in radiology, whereas gastrointestinal endoscopy remains comparatively underexplored and continues to rely predominantly on classification-oriented or template-based outputs \citep{min2021endoscopy,aslan2023colitisai}.

The emergence of large language models (LLMs) and multimodal foundation models has substantially expanded these capabilities. General-purpose architectures such as T5 \citep{raffel2020exploring} and LLaVA \citep{liu2023llava} demonstrated strong capabilities in natural-language generation, cross-modal alignment, and visual reasoning, stimulating their adaptation to medical applications including report generation, clinical documentation, and visual question answering. Within healthcare, domain-specific pretraining and multimodal adaptation have further improved visual--textual alignment. MedicalCLIP, MedCLIP, and BLIP-based medical adaptations \citep{yao2022medclip,chen2023blip}, together with more recent foundation models such as LLaVA-Med \citep{li2024llavamed}, BiomedGPT \citep{zhang2024biomedgpt}, Med-Flamingo \citep{moor2024medflamingo}, and Qwen2-VL \citep{bai2025qwen2vl}, have demonstrated strong capabilities in clinical language generation and cross-modal reasoning. These models leverage large-scale pretraining and instruction tuning to improve linguistic fluency and generalization, while parameter-efficient adaptation strategies such as LoRA \citep{hu2022lora} facilitate specialization to medical domains with reduced computational requirements.

Concurrently, increasing attention has been directed toward grounded medical vision--language generation, in which generated descriptions are explicitly linked to visual evidence rather than relying primarily on language priors. Stronger visual grounding has been shown to improve factual consistency, reduce unsupported clinical statements, and enhance clinical reliability \citep{singh2022explainability,yuan2023faithfulcap,wu2025groundedmedical}. This shift has also motivated evaluation protocols specifically targeting hallucination detection, factuality, and evidence attribution in medical vision--language models \citep{chen2025hallucination,wu2025groundedmedical}. Such considerations are particularly important in clinical narrative generation, where linguistically plausible but visually unsupported findings may lead to clinically misleading interpretations.

Despite their impressive generative capabilities, current foundation vision--language models frequently rely on global image representations or weakly supervised attention mechanisms \citep{li2024llavamed,zhang2024biomedgpt,moor2024medflamingo}, providing only partial correspondence between generated descriptions and localized pathological findings. This limitation is particularly relevant in gastrointestinal endoscopy, where clinical interpretation depends on subtle inflammatory findings and on their spatial distribution and co-occurrence. Moreover, computational cost, limited annotated data, and the need for interpretable reasoning remain important barriers to clinical deployment \citep{bai2025qwen2vl,hu2022lora}. Consequently, explicit lesion-level grounding, relational reasoning, and token-level attribution represent a complementary direction to increasing model scale. The proposed \method{} framework follows this direction by conditioning clinical language generation on localized lesion representations and their relationships, explicitly linking generated descriptions to the pathological evidence supporting them.

\subsection{Research Gaps and Clinical Motivation for Lesion-Aware Relational Grounding}

Despite substantial progress in medical vision--language modeling, current frameworks still lack unified approaches capable of jointly integrating lesion localization, disease severity assessment, and explainable caption generation. Most existing systems remain focused either on classification accuracy or isolated captioning pipelines without explicitly linking localized pathological evidence to clinical narratives. This limitation is compounded by the scarcity of dual-annotated datasets containing both lesion-level annotations and expert-authored textual descriptions, as annotation costs, inter-observer variability, and limited availability of clinical narratives continue to constrain the development of grounded medical captioning systems \citep{li2020comparison,li2021skinreport,yan2022retina}. Although the UC-Caption dataset introduced by \citet{lopezescamilla2025lesionaware} partially addresses this limitation, large-scale multimodal supervision remains limited in ulcerative colitis imaging.

These challenges are particularly relevant in gastrointestinal endoscopy, where clinical interpretation is inherently evidence-driven and lesion-centric. During endoscopic assessment, gastroenterologists identify specific pathological findings such as vascular attenuation, erythema, friability, erosions, ulceration, and bleeding, and subsequently integrate their presence, spatial distribution, and co-occurrence into an assessment of disease severity. Consequently, a computational framework intended to support clinical interpretation should not only recognize individual pathological findings, but also preserve their spatial and contextual relationships.

Most existing medical vision--language models generate descriptions from global visual representations that compress the entire image into a single latent embedding. Although effective for image-level prediction and generic caption generation, this strategy can obscure the contribution of individual pathological findings and weaken the correspondence between generated statements and observable visual evidence. As a result, language generation may become influenced by learned linguistic priors rather than by directly identifiable pathological structures, increasing the risk of unsupported findings, factual inconsistencies, and clinically misleading descriptions.

Lesion-level grounding partially addresses this limitation by establishing explicit correspondence between generated language and localized visual evidence. However, localization alone is insufficient to fully represent clinical reasoning in ulcerative colitis, where disease severity often emerges from the co-occurrence, distribution, and interaction of multiple inflammatory findings rather than from isolated lesions. Graph-based representations provide a natural mechanism for modeling these dependencies by representing pathological findings as nodes and their spatial or semantic relationships as edges, thereby preserving lesion-level information while enabling relational reasoning across the endoscopic scene.

These considerations motivate the development of lightweight, interpretable, and lesion-aware multimodal frameworks in which localized pathological evidence and inter-lesion relationships directly influence language generation. The proposed \method{} framework follows this direction by combining lesion-level grounding with graph-conditioned reasoning, enabling generated clinical descriptions to remain explicitly connected to both the location of pathological findings and the relationships among them.

\begin{figure*}[t]
    \centering
    \includegraphics[width=\textwidth]{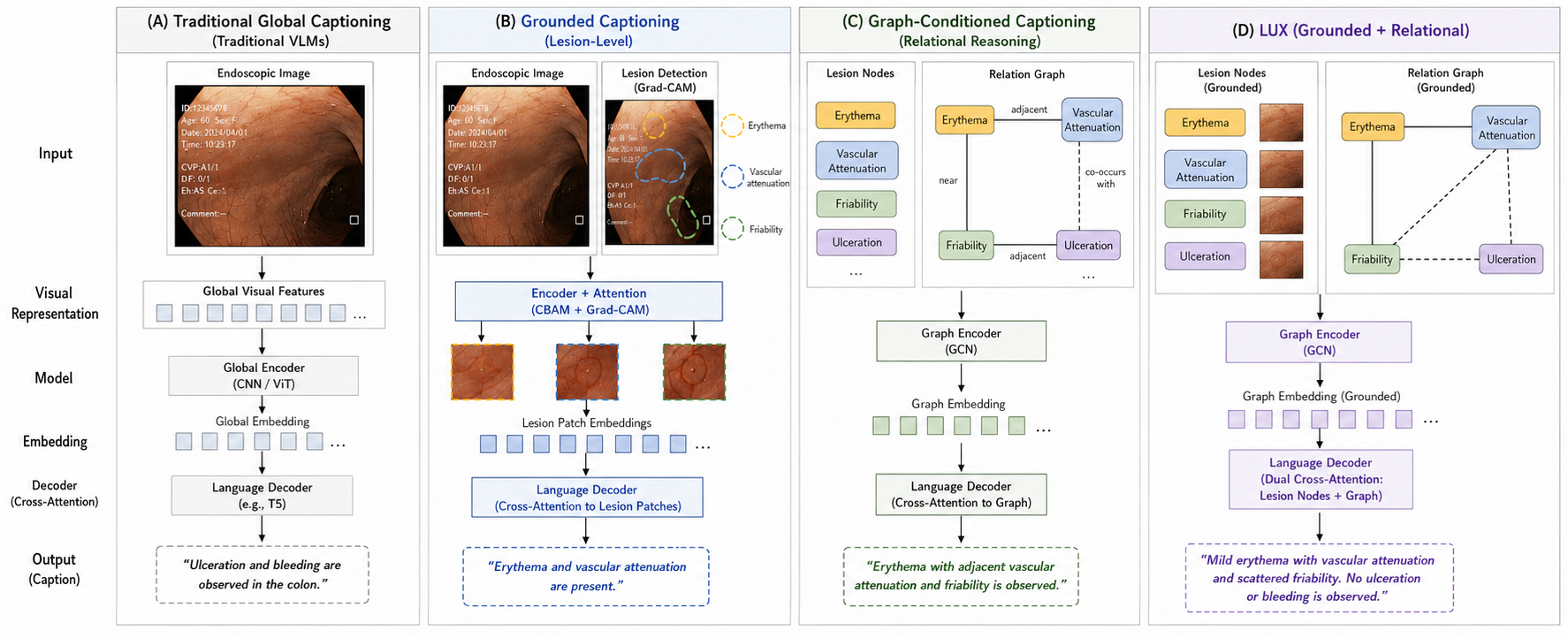}
    \caption{
Evolution of medical vision--language captioning paradigms. 
Traditional approaches rely on global visual embeddings, whereas grounded methods associate generated descriptions with localized lesion evidence. 
Graph-conditioned architectures additionally capture spatial and semantic interactions among pathological findings. 
The proposed LUX framework unifies lesion-level grounding and graph-based relational reasoning through dual cross-attention, producing clinically interpretable captions supported by explicit pathological evidence.
}
    \label{fig:grounding_relational_reasoning}
\end{figure*}
\subsection{Mayo Endoscopic Subscore and the Need for Explainable Multimodal Reasoning}
\label{sec:identified_gap_mes}

The \textit{Mayo Endoscopic Subscore (MES)} is one of the most widely used clinical scales for assessing disease activity in ulcerative colitis during colonoscopy. MES grades range from 0 (remission) to 3 (severe disease) and are based on qualitative interpretation of mucosal findings including vascular attenuation, erythema, friability, erosions, and spontaneous bleeding. Although widely adopted in clinical practice and therapeutic monitoring, MES evaluation remains inherently subjective, particularly in borderline cases where multiple inflammatory features co-occur.

Table~\ref{tab:mes_definition} summarizes the MES grading scale and its dominant endoscopic characteristics.

\begin{table}[t]
\centering
\small
\setlength{\tabcolsep}{6pt}
\renewcommand{\arraystretch}{1.25}
\begin{tabularx}{\linewidth}{c >{\raggedright\arraybackslash}p{3.2cm} >{\raggedright\arraybackslash}X}
\toprule
\textbf{MES} & \textbf{Clinical Description} & \textbf{Dominant Endoscopic Characteristics} \\
\midrule
0 &
Normal or inactive disease &
Normal mucosa, intact vascular pattern, no friability, absence of bleeding \\
\midrule
1 &
Mild disease &
Mild erythema, decreased vascular pattern visibility, minimal granularity, no spontaneous bleeding \\
\midrule
2 &
Moderate disease &
Marked erythema, absent vascular pattern, friability, erosions, contact bleeding \\
\midrule
3 &
Severe disease &
Spontaneous bleeding, ulceration, extensive friability, mucosal destruction \\
\bottomrule
\end{tabularx}
\caption{Mayo Endoscopic Subscore (MES) definitions and dominant endoscopic characteristics used for ulcerative colitis severity assessment.}
\label{tab:mes_definition}
\end{table}

A major limitation of current computational approaches is that MES prediction is frequently learned from global image embeddings without preserving explicit correspondence between localized pathological findings and the resulting clinical interpretation. Consequently, many systems can predict severity scores while failing to explain which lesions contributed to the decision or how visual evidence supports the generated description.

\begin{figure*}[t]
    \centering
    \includegraphics[width=0.8\linewidth]{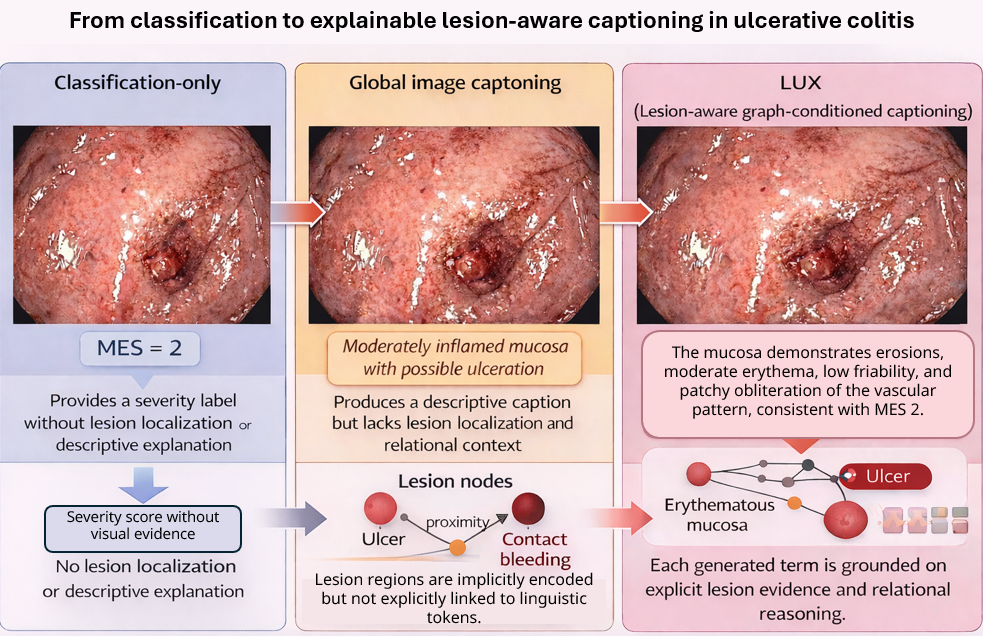}
    \caption{
    Conceptual comparison between image-level classification, global image captioning, and the proposed lesion-aware visual--language framework.
    Image-level classification produces a single severity label without visual or linguistic justification.
    Global image captioning generates descriptive text but lacks explicit lesion localization and relational context.
    In contrast, the proposed \method{} framework constrains language generation through lesion-centric representations and graph-conditioned reasoning, ensuring that each generated term is grounded on explicit lesion evidence and inter-lesion relationships.
    }
    \label{fig:evolution}
\end{figure*}

This limitation is particularly evident in medical captioning systems that generate fluent but weakly grounded clinical narratives, baseline models may hallucinate inflammatory findings or misestimate disease severity despite the absence of supporting visual evidence. As conceptually summarized in Fig.~\ref{fig:evolution}, current paradigms progressively transition from image-level prediction toward visually grounded language generation, although explicit lesion-centric reasoning remains largely absent.


These challenges motivate the need for multimodal frameworks capable of explicitly linking lesion-level evidence with structured clinical language. By conditioning language generation on lesion-aware relational representations, captioning systems can improve interpretability, reduce hallucinated findings, and provide clinically traceable reasoning aligned with endoscopic assessment practices.

The proposed \method{} framework addresses this limitation by integrating lesion localization, relational graph modeling, and graph-conditioned caption generation within a unified evidence-grounded pipeline.
\section{Proposed Approach}
\label{sec:method}

Building on the clinical and methodological motivation established in Section 2, LUX formulates endoscopic image captioning as an evidence-grounded multimodal reasoning process rather than as direct image-to-text translation.

The framework follows a progressive reasoning pipeline in which visually salient mucosal cues are first identified and encoded, subsequently aggregated into lesion-level entities, structured through spatial and semantic relationships, and finally verbalized as clinically grounded language. In this way, each processing stage constrains the next, reducing the dependence of caption generation on global visual representations or linguistic priors alone.

As illustrated in Fig. 2, LUX comprises five interacting components: (i) a hybrid visual encoder, (ii) a scene–lesion graph construction module, (iii) a lesion-aware captioning decoder, (iv) a linguistic formatting and structural generation module, and (v) an explainability component.

An input endoscopic image is first processed through a CBAM-enhanced ResNet-50 visual encoder designed to preserve multiscale mucosal information while selectively emphasizing clinically relevant regions. The resulting attention-refined visual representations provide the basis for lesion localization and lesion-level feature extraction.

Localized pathological evidence is subsequently transformed into a structured scene–lesion graph. Each graph node represents a candidate pathological finding, whereas graph edges encode spatial proximity and semantic similarity among lesions. Graph convolutional propagation then integrates contextual information across related findings, producing lesion-aware relational representations that capture the organization of inflammatory evidence within the endoscopic scene.

The visual and graph representations jointly condition a T5 decoder through dual cross-attention. During token generation, the decoder dynamically integrates global visual context with lesion-level relational information, enabling generated clinical terms to remain associated with specific pathological evidence. Token-to-lesion alignment further provides explicit traceability between linguistic elements and lesion representations.

Following decoding, a lightweight deterministic linguistic formatting module standardizes clinical terminology, grammatical structure, and evidence ordering without modifying the underlying visual or relational reasoning. Finally, pixel-level Grad-CAM explanations and token-level lesion attribution provide complementary mechanisms for visualizing the evidence supporting the generated description.

The overall conceptual flow can therefore be summarized as:

visual evidence → lesion entities → relational graph → grounded language → explainable output.

The following sections describe each component in the order in which information propagates through the architecture.

\begin{figure*}[t]
    \centering
    \includegraphics[width=0.6\textwidth]{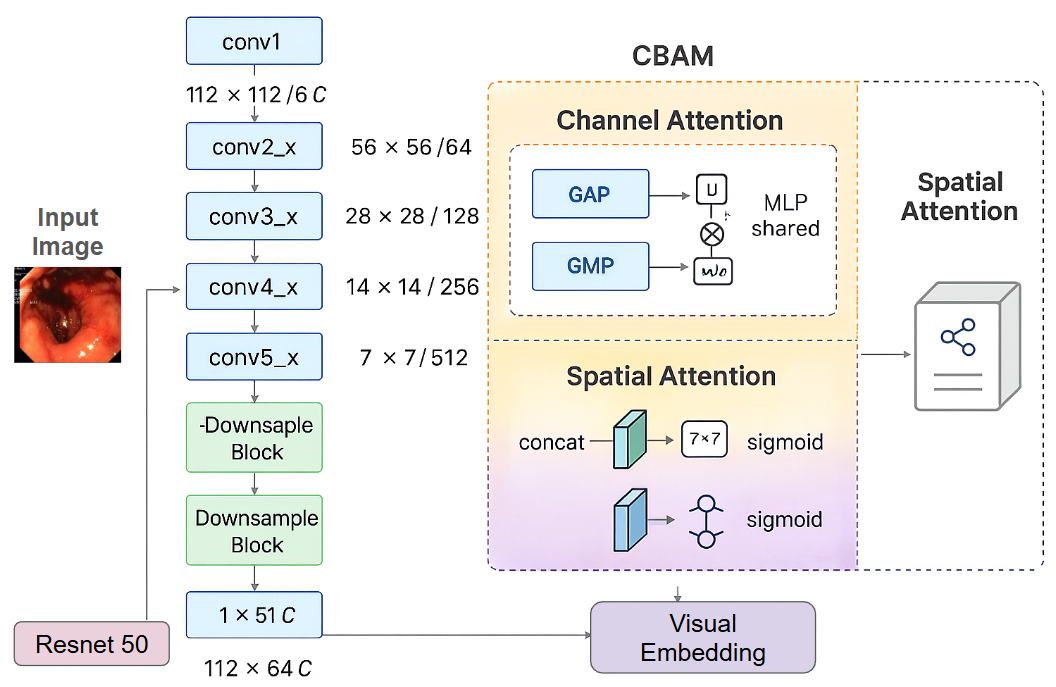}
    \caption{
        Detailed architecture of the CBAM--enhanced ResNet-50 visual encoder used in \method{}.
        The left side illustrates the full ResNet-50 convolutional hierarchy (\texttt{conv1} through \texttt{conv5\_x}), including downsampling blocks and the final spatial feature map projection.
        On the right, the Convolutional Block Attention Module (CBAM) is decomposed into its two sequential components: (1) \emph{Channel Attention}, which refines feature maps using global average pooling (GAP), global max pooling (GMP), and a shared MLP; and (2) \emph{Spatial Attention}, which aggregates feature cues via a $7{\times}7$ convolution followed by a sigmoid mask.
        The resulting attention-refined feature map constitutes the \textit{visual embedding} used for lesion--scene graph construction and T5 cross-attention decoding.
    }
    \label{fig:cbam_resnet}
\end{figure*}

\subsection{Component I: Visual Hybrid Encoder (ResNet-50 + CBAM)}
\label{sec:visual_encoder}

Endoscopic severity cues in ulcerative colitis are often subtle and spatially heterogeneous, including fine vascular attenuation, diffuse erythema, and granular texture changes. Capturing these cues requires a visual encoder capable of preserving spatial detail while selectively emphasizing clinically salient regions across multiple scales.

\textbf{ResNet-50 Backbone Architecture.}
The visual encoder of \method{} builds upon a ResNet-50 backbone initialized with ImageNet weights and adapted for colonoscopic imaging. All convolutional layers up to the \texttt{conv4\_x} block were retained, while the final residual block was modified to output $16\times16$ feature maps with 2048 channels. The hybrid visual encoding stage integrates CBAM-based attention with a T5-compatible representation, as illustrated in Fig.~\ref{fig:cbam_resnet}.

\textbf{CBAM Integration.}
To refine visual representations toward clinically relevant evidence, a Convolutional Block Attention Module (CBAM) is applied after each residual stage:
\begin{equation}
M_{\text{cbam}}(F) =
\sigma\big(W_s *
[\mathrm{AvgPool}(F_c);
\mathrm{MaxPool}(F_c)]\big)
\odot F_c
\label{eq:cbam}
\end{equation}
Here, $F_c$ denotes the channel-refined feature map, $\mathrm{AvgPool}(\cdot)$ and $\mathrm{MaxPool}(\cdot)$ represent global pooling operations, $W_s$ is a learnable spatial convolution kernel, $\sigma(\cdot)$ denotes the sigmoid activation, and $\odot$ indicates element-wise multiplication. This mechanism selectively amplifies discriminative mucosal patterns while suppressing irrelevant background information.

\begin{figure*}[t]
    \centering
    \includegraphics[width= 0.75\textwidth]{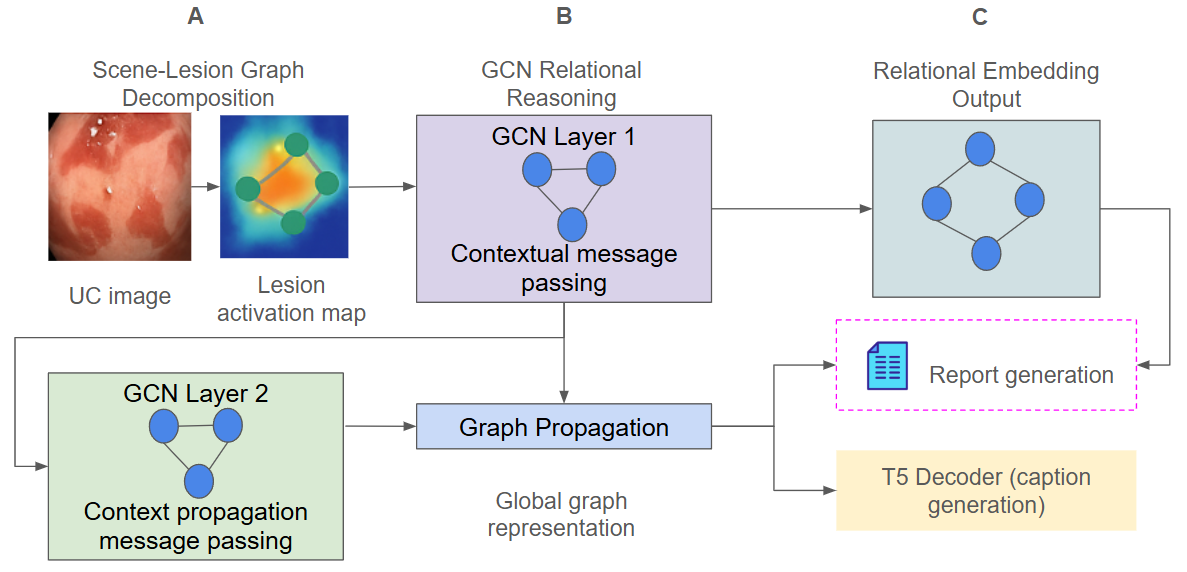}
    \caption{
    Scene--Lesion Graph decomposition and relational reasoning within the proposed \method{} framework.
    (A) Attention-refined activation maps derived from the CBAM-enhanced visual encoder are transformed into lesion-centric node embeddings, where each node represents a localized pathological finding.
    (B) Lesion nodes are connected through spatial proximity and semantic similarity relations, forming a structured scene--lesion graph that models interactions among inflammatory regions prior to reasoning.
    Stacked GCN layers propagate contextual information across connected lesions, enabling the model to capture relational inflammatory patterns rather than isolated visual cues.
    (C) The resulting graph-aware embeddings are injected into the T5 decoder through dual cross-attention, allowing generated clinical tokens to remain explicitly grounded in lesion-level pathological evidence.
    }
    \label{fig:scene_gcn}
\end{figure*}

\textbf{Multi-scale Fusion.}
Given that inflammatory cues may manifest at different spatial resolutions, features from \texttt{conv3\_x}, \texttt{conv4\_x}, and \texttt{conv5\_x} are unified:
\begin{equation}
F_{\text{fusion}} =
\mathrm{Conv}_{1\times1}
\big([F_3 \oplus F_4 \oplus F_5]\big)
\label{eq:fusion}
\end{equation}
In this formulation, $F_3$, $F_4$, and $F_5$ denote feature maps extracted at increasing receptive-field scales, $\oplus$ represents channel-wise concatenation, and the $1\times1$ convolution projects the concatenated features into a unified representation that preserves both fine-grained texture and high-level semantic context.

\textbf{Embedding Projection.}
The fused visual features are projected to $d_v = 768$ to ensure dimensional compatibility with the T5 decoder during cross-modal alignment.

\textbf{Spatial Attention Visualization.}
CBAM activations consistently highlight intact vascular structures in remission cases and ulcerated or bleeding regions in higher MES grades, providing structured visual priors for subsequent lesion modeling. 

Importantly, the architecture illustrated in Fig.~\ref{fig:cbam_resnet} does not operate as a generic feature extractor. Instead, the sequential combination of channel and spatial attention progressively constrains the visual representation toward clinically relevant inflammatory evidence. Channel attention emphasizes discriminative mucosal patterns associated with vascular attenuation, ulceration, and bleeding, whereas spatial attention preserves the localization structure required for lesion-level reasoning. This progressive refinement is essential for the subsequent graph-construction stage, since lesion nodes and relational embeddings depend directly on the spatial consistency of these attention-refined representations.

While these attention-refined feature maps indicate \emph{where} salient evidence may lie, they do not yet explicitly represent \emph{what} lesion entities exist or \emph{how} they relate to one another, motivating the next stage of the pipeline. To illustrate these conceptual differences, Fig.~\ref{fig:evolution} summarizes the progression from image-level classification and global medical captioning toward the proposed lesion-aware graph-conditioned reasoning framework across representative MES severity levels.

\subsection{Component II: Scene--Lesion Graph Construction}
\label{sec:graph}

\begin{figure*}[t]
    \centering
    \includegraphics[width=0.72\textwidth]{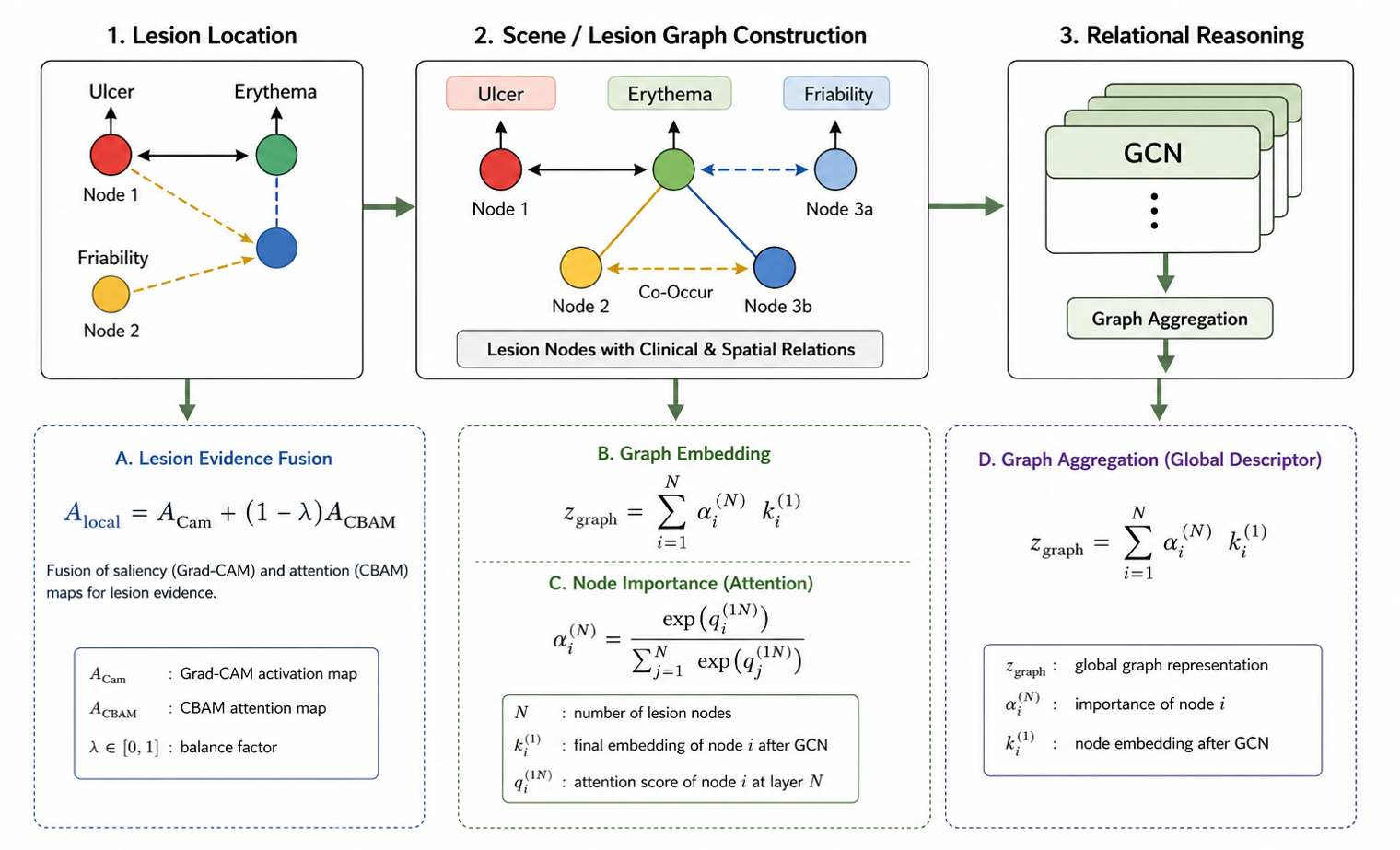}
    \caption{
    \textbf{Component II: Scene--Lesion Graph Construction.}
    Localized lesion evidence is transformed into a structured scene--lesion graph,
    where nodes represent pathological findings and edges encode spatial and
    clinical relations among lesions. This relational representation enables
    explicit lesion-centric reasoning prior to language generation.
    }
    \label{fig:component2}
\end{figure*}

Clinical reasoning operates at the level of pathological entities rather than raw pixel intensities. To enable lesion-centric and relational reasoning,
attention-derived visual evidence is converted into a structured scene--lesion graph. As illustrated in Fig.~\ref{fig:evolution}, the proposed framework therefore transitions from holistic image representations toward structured lesion-centric reasoning through explicit graph construction.
Figure~\ref{fig:component2} presents the complete lesion-graph construction pipeline, including lesion extraction, node formation, and relational graph assembly, whereas Fig.~\ref{fig:scene_gcn} focuses specifically on graph convolutional propagation and relational reasoning across lesion nodes. Conceptually, Fig.~\ref{fig:scene_gcn} illustrates the transition from localized visual evidence toward structured relational reasoning. This relational transition directly extends the conceptual progression previously summarized in Fig.~\ref{fig:evolution}, where lesion entities become the fundamental reasoning units driving caption generation.

Rather than treating inflammatory regions as isolated activations, the proposed framework organizes lesion representations into an explicit graph structure where pathological entities interact through spatial and semantic relations.
This formulation enables the model to reason over contextual lesion configurations, such as co-occurring ulceration, vascular attenuation, and bleeding patterns, before conditioning language generation.
Consequently, the generated captions are influenced not only by the presence of individual lesions, but also by their relational organization within the endoscopic scene.

\textbf{Grad-CAM Lesion Activation.}
Lesion candidate regions are localized using Grad-CAM:
\begin{equation}
L^c_{\text{GradCAM}} =
\mathrm{ReLU}
\left(
\sum_k \alpha_k^c A^k
\right)
\label{eq:gradcam}
\end{equation}
Here, $A^k$ denotes the activation map of the $k$-th convolutional channel, $\alpha_k^c$ represents the gradient-based importance weight for class $c$, and $\mathrm{ReLU}(\cdot)$ ensures that only positive contributions associated with lesion evidence are retained. This localization step corresponds to the lesion activation stage shown in Fig.~\ref{fig:component2}.

\textbf{Supervised Lesion Localization using Manual Annotations.}

To provide explicit spatial supervision during training, lesion localization is regularized using manually annotated lesion masks generated for the subset of 400 LIMUC images described in Section~4.2. These annotations provide the ground-truth lesion regions required to guide the localization branch while preserving the remaining LIMUC images for large-scale visual representation learning.

\begin{equation}
\mathcal{L}_{\text{loc}} =
1 -
\frac{
2|M_{\text{pred}}\cap M_{\text{gt}}|
}{
|M_{\text{pred}}|+|M_{\text{gt}}|
}
\label{eq:dice}
\end{equation}

In this Dice-based formulation, $M_{\text{pred}}$ denotes the predicted lesion activation mask and $M_{\text{gt}}$ corresponds to the manually annotated lesion mask associated with each training image. Minimizing this objective encourages spatial alignment between Grad-CAM-derived lesion activations and expert-defined pathological regions, thereby improving the anatomical consistency of the learned lesion representations.

\textbf{Node Embeddings.}
Each localized lesion region is abstracted into a graph node:
\begin{equation}
x_i=
\frac{1}{|R_i|}
\sum_{(u,v)\in R_i}
F_{\text{fusion}}(u,v)
L_{\text{GradCAM}}(u,v)
\label{eq:node_embedding}
\end{equation}
Here, $R_i$ represents the spatial region associated with lesion $i$, $F_{\text{fusion}}(u,v)$ is the fused visual feature at pixel location $(u,v)$, and $L_{\text{GradCAM}}(u,v)$ acts as a lesion relevance weight, yielding a lesion-centric embedding. Each embedding $x_i$ corresponds to a lesion node illustrated in Fig.~\ref{fig:scene_gcn}.

\textbf{Edge Weights.}
Edges encode both spatial proximity and semantic similarity:
\begin{equation}
w_{ij}=
\lambda_s
e^{-\|p_i-p_j\|^2/\sigma_s^2}
+
\lambda_c
\frac{x_i\cdot x_j}
{\|x_i\|\|x_j\|}
\label{eq:edge_weights}
\end{equation}
In this expression, $p_i$ and $p_j$ denote the spatial centroids of lesions $i$ and $j$, respectively; the first term captures spatial adjacency, while the second encodes semantic similarity between lesion embeddings. The coefficients $\lambda_s$ and $\lambda_c$ balance spatial and semantic contributions. These edge weights define the graph connectivity illustrated in Fig.~\ref{fig:scene_gcn}, where spatial adjacency and semantic affinity jointly determine relational connectivity between lesion entities.

The coefficients $\lambda_s$ and $\lambda_c$ balance spatial proximity and semantic similarity during graph construction. Since no prior clinical assumption favors either spatial adjacency or feature similarity as dominant factors for lesion interaction, both terms are assigned equal weight.

Accordingly, we set $\lambda_s = 0.5$ and $\lambda_c = 0.5$, representing a neutral prior that avoids biasing graph connectivity toward either geometric or feature-based relations. Preliminary validation experiments indicated that this symmetric configuration provided stable graph formation and consistent training convergence. Learning these coefficients was intentionally avoided to preserve interpretability of lesion relations.

\textbf{GCN Propagation.}
Relational reasoning is performed using stacked GCN layers:
\begin{equation}
H^{(l+1)}=
\sigma(\hat{A}H^{(l)}W^{(l)}),
\quad H^{(0)}=X
\label{eq:gcn}
\end{equation}
Here, $\hat{A}$ denotes the normalized adjacency matrix of the lesion graph, $W^{(l)}$ is the learnable weight matrix at layer $l$, and $\sigma(\cdot)$ is a nonlinear activation function enabling information propagation across related lesion nodes.

\textbf{Global Graph Descriptor.}
A global graph representation is obtained via attention pooling:
\begin{equation}
z_{\text{graph}}=
\sum_i \alpha_i z_i
\label{eq:graph_pool}
\end{equation}
In this formulation, $z_i$ denotes the embedding of lesion node $i$ and $\alpha_i$ represents its learned importance weight, producing a compact summary of the lesion configuration. The resulting graph-level descriptor corresponds to the aggregation stage shown in Fig.~\ref{fig:scene_gcn}.

The construction and propagation of lesion-level relations through graph convolutional layers are illustrated in Fig.~\ref{fig:scene_gcn}, which focuses specifically on relational reasoning within the scene--lesion graph prior to language generation. 

At this stage, localized pathological evidence has been transformed into an explicit relational representation in which lesion interactions, spatial organization, and semantic affinity collectively constrain downstream language generation.

\begin{figure*}[t]
    \centering
    \includegraphics[width=.952\textwidth]{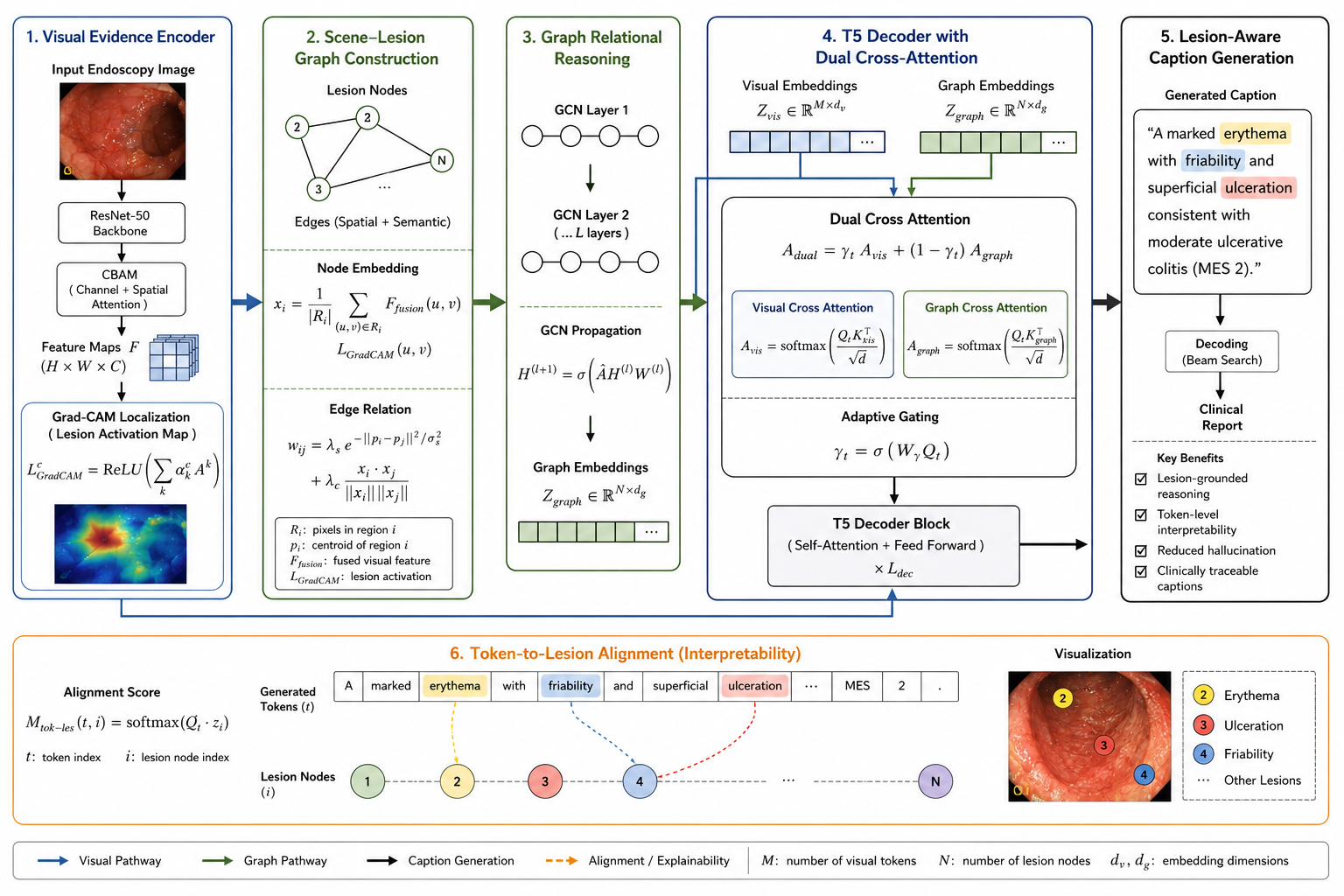}
    \caption{
    \textbf{Component III: Lesion-Aware Captioning.}
    The scene--lesion graph conditions the T5 decoder through dual cross-attention,
    allowing generated tokens to attend jointly to visual embeddings and structured
    lesion relations, thereby grounding clinical captions in localized pathological evidence.
    }
    \label{fig:component3}
\end{figure*}

\subsection{Component III: Lesion-Aware Captioning (T5 Decoder + Dual Cross-Attention)}
\label{sec:captioning}

\begin{figure}[t]
    \centering
    \includegraphics[width= 0.45\textwidth]{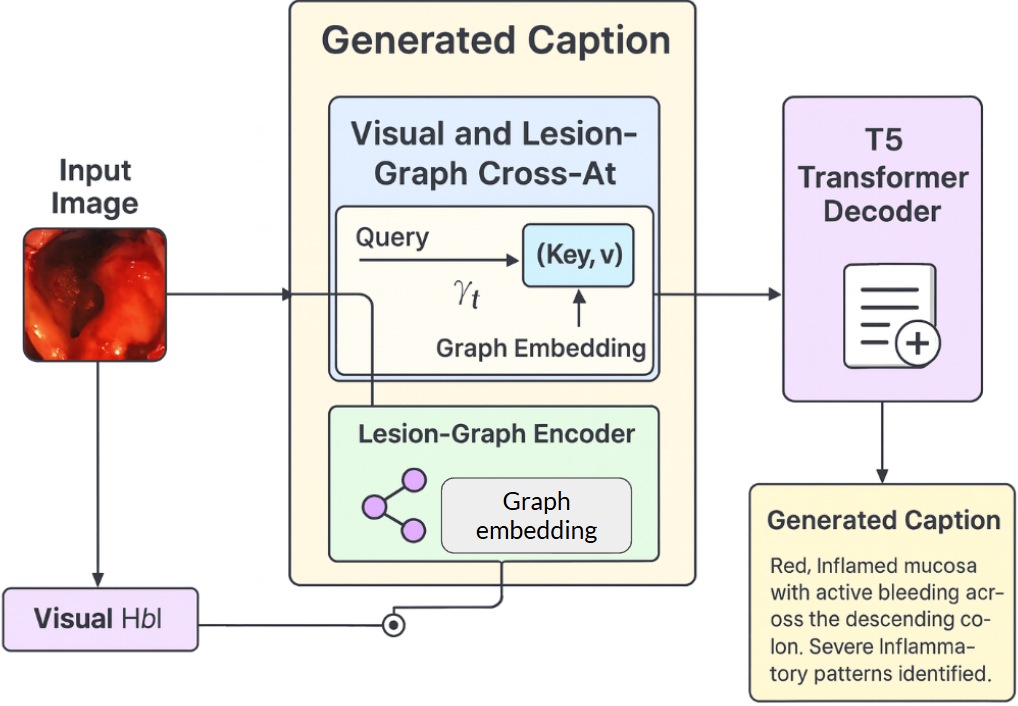}
    \caption{\textbf{Lesion-aware captioning module (Component III).}
    The T5 decoder receives two complementary information streams:
    (1) dense visual embeddings extracted from the CBAM--ResNet encoder, and
    (2) structured relational embeddings produced by the lesion--scene graph.
    Each decoder query attends jointly to both sources through dual cross-attention.
    A learnable gating parameter $\gamma_t$ adaptively balances visual context and
    graph-derived lesion semantics at each decoding step, enabling the model to
    generate clinically grounded descriptions. The relational embedding aligns each
    generated word with the most relevant lesion node, supporting traceable and
    explainable caption generation.}
    \label{fig:lesion_caption}
\end{figure}

Once lesion evidence is encoded as a graph, the language model must be prevented from reverting to generic but weakly grounded descriptions. To achieve this, visual and graph embeddings are injected directly into the decoding process. This transition from descriptive captioning toward graph-conditioned language generation corresponds to the final reasoning stage illustrated in Fig.~\ref{fig:evolution}.
Figure~\ref{fig:component3} illustrates the complete graph-conditioned captioning pipeline within the global LUX architecture, including visual embeddings, lesion-graph conditioning, and token generation. In contrast, Fig.~\ref{fig:lesion_caption} focuses exclusively on the internal decoder mechanism, detailing how dual cross-attention and adaptive gating perform token-to-lesion grounding during caption generation.

\textbf{Dual Cross-Attention.}
Each decoder token attends jointly to visual features and lesion graph embeddings:
\begin{equation}
A_{\text{dual}}=
\gamma_t A_{\text{vis}}
+
(1-\gamma_t)A_{\text{graph}},
\quad
\gamma_t=\sigma(W_\gamma Q_t)
\label{eq:dual_attention}
\end{equation}
Here, $A_{\text{vis}}$ and $A_{\text{graph}}$ denote attention maps over visual features and lesion graph embeddings, respectively; $Q_t$ is the decoder query at time step $t$; $W_\gamma$ is a learnable projection matrix; and $\gamma_t$ dynamically balances visual context and lesion-centric reasoning during token generation. The lesion-aware captioning mechanism and its interaction with the language decoder are depicted in Fig.~\ref{fig:lesion_caption}.

\textbf{Token-to-Lesion Alignment.}
Token-level grounding is enforced via:
\begin{equation}
M_{\text{tok-les}}(t,i)
=
\mathrm{softmax}(Q_t\cdot z_i)
\label{eq:token_lesion}
\end{equation}
This alignment matrix quantifies the contribution of lesion node $i$ to the generation of token $t$, enabling explicit traceability between linguistic elements and pathological evidence. An overview of the graph-conditioned captioning stage is provided in Fig.~\ref{fig:component3}.

\subsection{Component IV: Linguistic Formatting and Structural Generation}
\label{sec:formatting}

Although lesion-aware decoding enforces visual grounding during caption generation, raw outputs produced by the language decoder may still exhibit linguistic variability, redundant phrasing, or terminology inconsistent with clinical reporting standards. In routine endoscopic practice, descriptions follow relatively stable narrative conventions aligned with disease severity assessment and structured documentation.

To ensure clinical usability and reporting consistency, a lightweight linguistic formatting module is introduced as a post-decoding stage. This component operates deterministically and does not modify visual or lesion-level reasoning, but standardizes textual realization while preserving semantic alignment with underlying pathological evidence.

\paragraph{Input representation.}
The module receives the token sequence generated by the T5 decoder:
\begin{equation}
S = \{w_1, w_2, \dots, w_T\}
\label{eq:decoder_sequence}
\end{equation}
representing a lesion-grounded descriptive caption prior to normalization.

\paragraph{Controlled clinical vocabulary alignment.}
Generated tokens are mapped to a curated clinical vocabulary derived from established ulcerative colitis descriptors (e.g., vascular attenuation, erythema, friability, erosion, bleeding). Synonymous or linguistically variable expressions are replaced using dictionary-based normalization rules to reduce semantic variability across captions.

\paragraph{Grammatical normalization.}
Rule-based transformations are applied to enforce syntactic consistency, including removal of redundant modifiers, correction of tense inconsistencies, and normalization of anatomical references. These operations rely on regular-expression parsing and predefined grammatical templates rather than additional neural training.

\paragraph{Structural sentence organization.}
To align generated captions with clinical documentation practices, findings are reordered following an evidence-first reporting structure:
(i) mucosal appearance,
(ii) vascular pattern,
(iii) presence of lesions,
(iv) bleeding or ulceration indicators.
This ordering mirrors expert endoscopic narrative reasoning and improves interpretability during clinical review.

\paragraph{Output representation.}
The resulting formatted caption preserves lesion-grounded semantics while providing a standardized and clinically interpretable description suitable for comparison across cases and severity grades.

Importantly, this module introduces no trainable parameters and therefore maintains full transparency and reproducibility of the caption generation pipeline.

\subsection{Component V: Explainability (Grad-CAM + Lesion Attribution)}
\label{sec:explainability}

Finally, clinical deployment requires not only accurate descriptions but also traceability of evidence. \method{} therefore provides explanations at complementary resolutions.

\textbf{(1) Pixel-level visual explanations} are obtained via Grad-CAM heatmaps aligned with CBAM activations.

\textbf{(2) Token-level lesion attribution} is derived from $M_{\text{tok-les}}$, indicating which lesion node influenced each generated token.

Together, these mechanisms ensure that explainability is an intrinsic property of the pipeline, enabling transparent alignment between visual evidence, lesion reasoning, and clinical language.
\section{Datasets and Clinical Context}
\label{sec:datasets}

Having defined the full evidence-to-language workflow, we now describe the datasets and clinical context supporting each stage of \method{}, from MES supervision and lesion localization priors to expert-authored descriptions required for grounded caption learning.

Two complementary datasets were employed to develop and evaluate the proposed \method{} framework: the large-scale \textbf{LIMUC (Lesion Image dataset for Ulcerative Colitis)} in \citet{limuc2022} and the \textbf{UC-Caption} dataset curated for multimodal captioning in \citet{lopezescamilla2025lesionaware}.  
Both datasets consist of anonymized colonoscopic frames labeled according to the \textit{Mayo Endoscopic Subscore (MES)}, ranging from 0 (remission) to 3 (severe disease), reflecting the visual severity of inflammation under standard white-light endoscopy.  
All studies were conducted retrospectively under institutional review board approval, ensuring compliance with ethical and privacy regulations.

\begin{figure}[t]
    \centering
    \includegraphics[width= 0.5\textwidth]{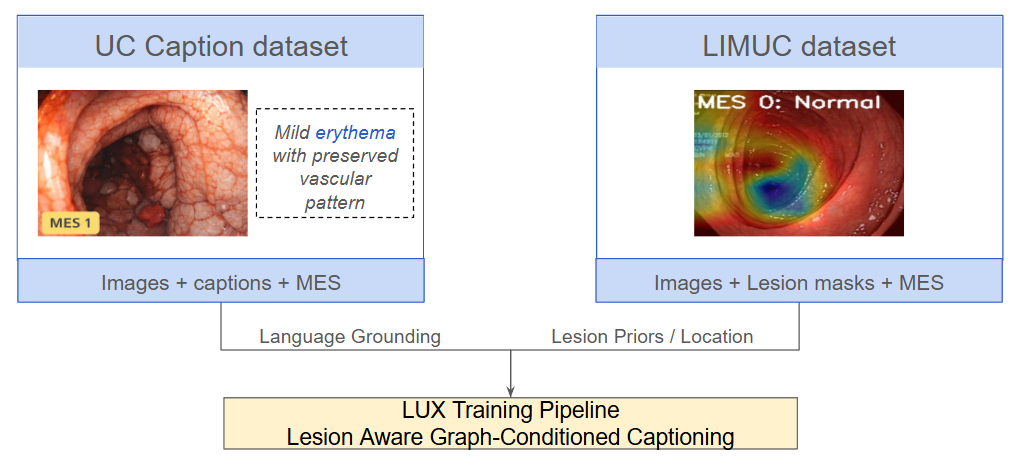}
    \caption{\textbf{Dataset methodology and integration strategy.}
    UC-Caption is a captioned dataset providing paired endoscopic images, expert-authored clinical descriptions, and MES labels, enabling supervised learning of clinically grounded language. LIMUC is a non-captioned auxiliary dataset providing endoscopic images and MES labels. Since the public LIMUC dataset does not include lesion localization masks, a manually annotated subset of 400 images was created to supervise the localization branch during training, while the remaining images were used for encoder pretraining, lesion priors, and lesion-centric relational learning. Together, both datasets provide complementary supervision that couples linguistic grounding (UC-Caption) with large-scale lesion-aware visual evidence (LIMUC) within the proposed training pipeline.}
    \label{fig:dataset_methodology}
\end{figure}

To support clinically grounded caption learning while preserving lesion-level supervision at scale, we adopt a dual-dataset methodology in which each dataset provides a distinct supervisory signal. UC-Caption supplies paired image--text examples defining the target clinical narrative, whereas LIMUC contributes large-scale visual diversity and MES supervision. A LUX-specific subset of LIMUC additionally provides manually created lesion-level annotations that strengthen localization and relational priors. Figure~\ref{fig:dataset_methodology} summarizes how both datasets are integrated within the training strategy.

\subsection{Primary Captioned Dataset (UC Caption baseline)}
\label{sec:uc_caption_dataset}

The primary dataset, previously introduced in \cite{lopezescamilla2025lesionaware}, comprises 500 high-resolution white-light endoscopic frames captured from patients diagnosed with ulcerative colitis. The dataset was constructed through a multistage curation and clinical annotation process designed to preserve correspondence between visible mucosal evidence, disease severity, and the resulting textual description. It spans all four MES grades: MES~0 (remission), MES~1 (mild activity), MES~2 (moderate activity), and MES~3 (severe activity).

\paragraph{Image curation.}
Candidate frames were screened for technical quality, clinical relevance, and visual redundancy. Frames with substantial motion blur, insufficient illumination, excessive specular reflection, obstructed mucosal views, visible endoscopic instruments, or insufficiently informative tissue were excluded. Near-duplicate frames depicting the same mucosal region and comparable findings were also removed. The retained images were selected to represent the spectrum of inflammatory manifestations relevant to MES assessment, including vascular pattern loss, erythema, friability, erosions, ulceration, and bleeding.

\paragraph{Clinical annotation and MES reconciliation.}
Each retained frame was independently reviewed by three clinicians, who assigned an MES grade and identified the inflammatory findings directly observable in the image. Inter-rater reconciliation was applied to cases presenting disagreement in severity or finding interpretation, thereby improving descriptive consistency and diagnostic reliability, particularly for visually adjacent MES categories. This review produced the consensus clinical evidence used for caption construction.

\paragraph{Caption generation and quality control.}
Each image was paired with a descriptive caption authored from the reconciled clinical findings. The captions provide concise observations of mucosal appearance and follow a controlled linguistic template derived from the Ulcerative Colitis Endoscopic Index of Severity (UCEIS) descriptors. Caption construction emphasized vascular pattern, mucosal surface, erosions or ulceration, and bleeding, and was restricted to evidence visible in the corresponding frame. Terminology and sentence structure were subsequently standardized to reduce purely linguistic variation without suppressing clinically meaningful differences among cases. Each sample is represented as a triplet $(I, C, y_{\text{MES}})$, where $I$ denotes the endoscopic image, $C$ the corresponding expert-authored caption, and $y_{\text{MES}} \in \{0,1,2,3\}$ the reconciled disease grade. This dataset served as the primary supervision source for image--language alignment and caption generation. The subsequent caption preprocessing and semantic-diversification procedures were applied only after construction of this clinically reviewed base corpus and are described in Appendix~\ref{appendix:preprocessing}.

\subsection{Auxiliary Dataset: LIMUC}
\label{sec:limuc_dataset}

To increase robustness and enable cross-cohort generalization, the LIMUC dataset, introduced in \citet{limuc2022}, was integrated as a secondary non-captioned corpus. The public release contains 11,276 colonoscopy images obtained from 564 patients across 1,043 colonoscopy procedures performed between December 2011 and July 2019 at the Department of Gastroenterology of Marmara University School of Medicine. The images are organized into four Mayo Endoscopic Subscore categories and include 6,105 MES~0 images, 3,052 MES~1 images, 1,254 MES~2 images, and 865 MES~3 images. This distribution is strongly weighted toward remission and mild disease, while still providing representative variability in vascularity, mucosal granularity, ulceration, bleeding, illumination, and image appearance.

\paragraph{Original MES annotation protocol.}
All LIMUC images were blindly and independently classified by two experienced gastroenterologists. Images receiving discordant labels were subsequently reviewed by a third experienced gastroenterologist, who did not have access to the previous decisions, and the final MES label was determined by majority voting. The public release also provides patient-grouped images, enabling patient-level data partitioning and preventing images from the same patient from being distributed across training and evaluation subsets. Although LIMUC provides expert-assigned MES labels, it does not contain paired textual captions or pixel-level lesion localization masks. Its original annotations therefore support auxiliary MES training and visual representation learning but do not directly supervise language generation or lesion segmentation.

All images were anonymized at acquisition time, and no personally identifiable information was retained. The dataset was used strictly for methodological evaluation and model pretraining. Ethical approval was obtained for retrospective analysis, and all procedures adhered to the Declaration of Helsinki.

\paragraph{Construction of the LUX lesion-mask subset.}
To provide explicit lesion-level supervision for the proposed localization branch, a manually annotated subset of 400 LIMUC images was constructed. Image selection followed a stratified sampling strategy, with 100 images selected from each MES category (MES 0--3). Within each severity group, candidate images were randomly sampled and subsequently screened to retain representative cases exhibiting diverse inflammatory manifestations and visually challenging patterns that could potentially lead to ambiguity between adjacent MES grades. Each selected image was manually annotated at the pixel level by the first author. The resulting lesion masks delineate clinically relevant inflammatory findings, including vascular pattern alterations, erythema, erosions, ulceration, and bleeding whenever present. When multiple spatially disconnected regions represented the same inflammatory process, each visible region was retained in the corresponding mask; background mucosa, endoscopic borders, specular highlights, and non-pathological image artifacts were excluded. These annotations were created exclusively to supervise the localization loss described in Section~\ref{sec:graph} and were not used for caption generation or during inference. Consequently, the complete LIMUC dataset contributed to visual representation learning, whereas only the manually annotated subset provided lesion-level supervision during model optimization.

\begin{figure*}[t]
    \centering

    \begin{subfigure}{0.32\textwidth}
        \centering
        \includegraphics[width=\linewidth]{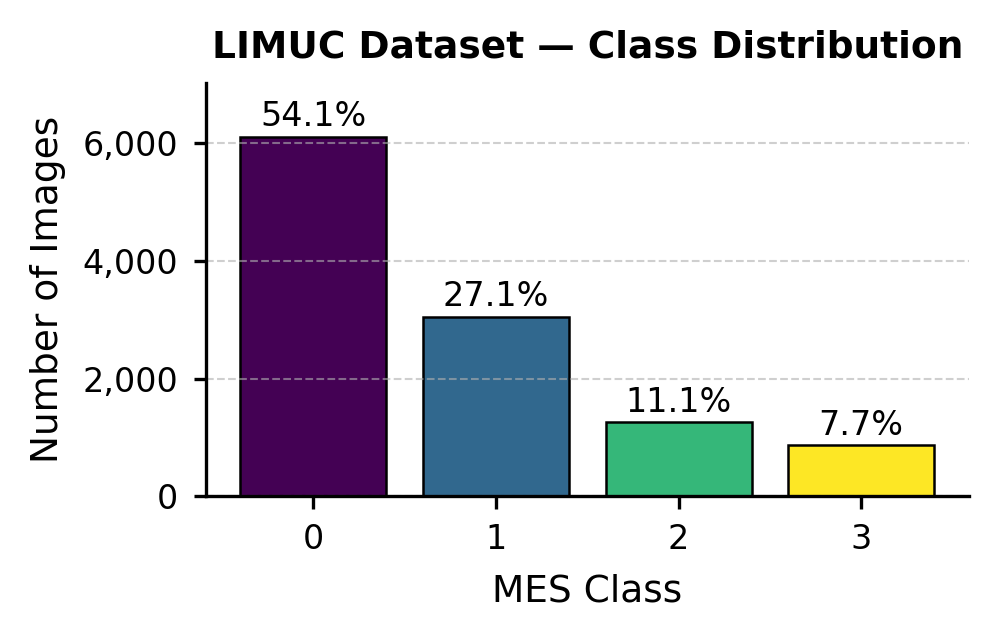}
        \caption{Class distribution of the LIMUC dataset (MES 0--3).}
        \label{fig:limuc_dist}
    \end{subfigure}
    \hfill
    \begin{subfigure}{0.32\textwidth}
        \centering
        \includegraphics[width=\linewidth]{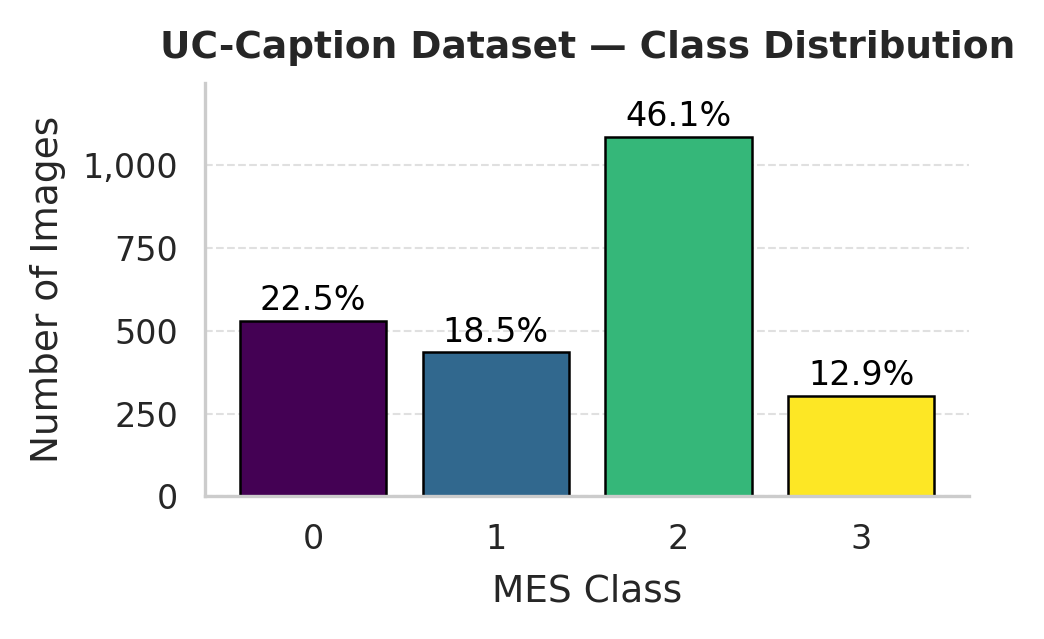}
        \caption{Class distribution of the UC-Caption dataset (MES 0--3).}
        \label{fig:uc_caption_dist}
    \end{subfigure}
    \hfill
    \begin{subfigure}{0.32\textwidth}
        \centering
        \includegraphics[width=\linewidth]{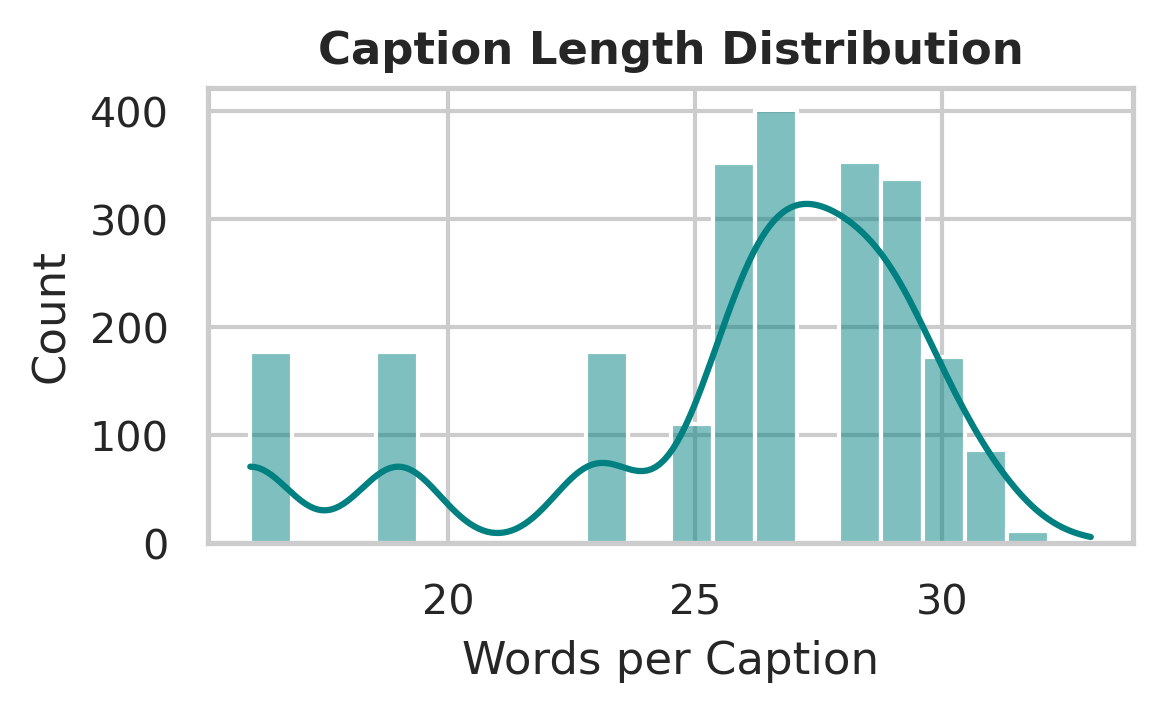}
        \caption{Caption length distribution (UC-Caption dataset).}
        \label{fig:uc_caption_length}
    \end{subfigure}

    \caption{Dataset statistics illustrating class imbalance and linguistic variability. 
    (A) The LIMUC dataset exhibits predominance of remission and mild cases (MES 0--1), providing large-scale visual diversity. 
    (B) The UC-Caption dataset shows a more balanced representation across moderate and severe classes, suitable for linguistic supervision. 
    (C) Caption length distribution indicates clinically structured yet lexically rich descriptions, with a mean of $\sim$25 words per caption.}
    \label{fig:dataset_distributions}
\end{figure*}

\begin{figure*}[t]
    \centering

    \begin{subfigure}{0.32\textwidth}
        \centering
        \includegraphics[width=\linewidth]{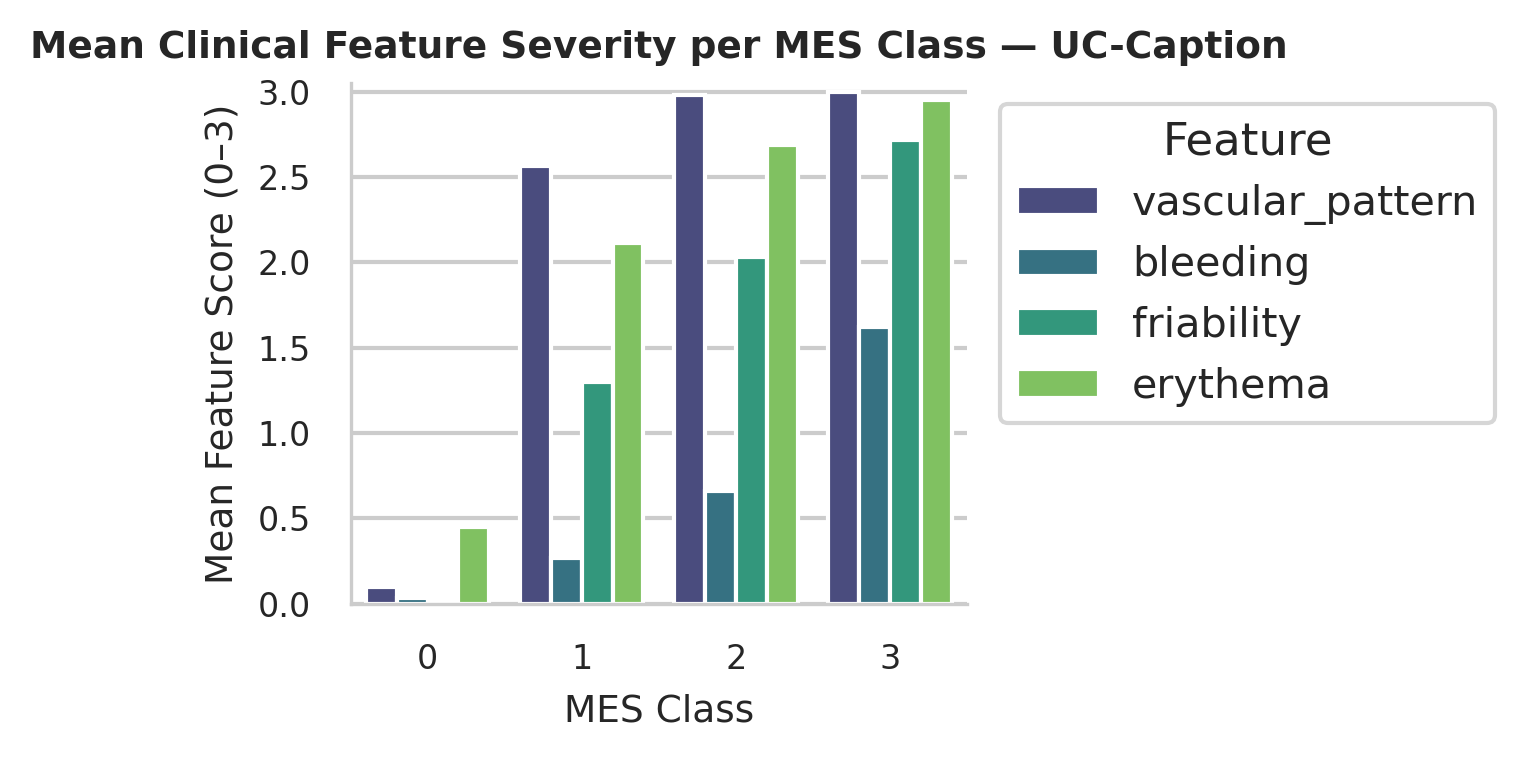}
        \caption{Mean clinical feature severity per MES class.}
        \label{fig:uc_feature_mean}
    \end{subfigure}
    \hfill
    \begin{subfigure}{0.32\textwidth}
        \centering
        \includegraphics[width=\linewidth]{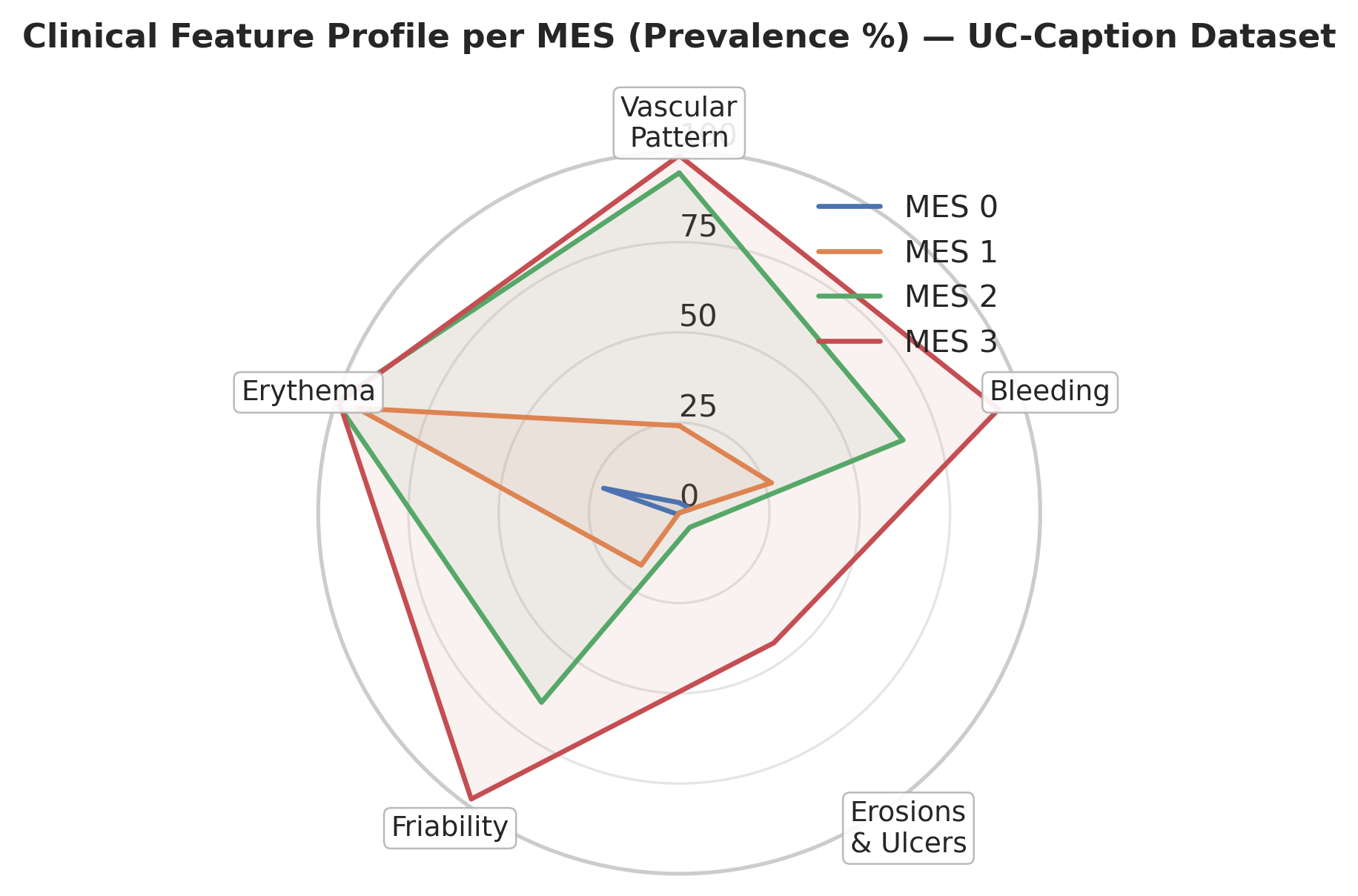}
        \caption{Clinical feature profile per MES (radar plot).}
        \label{fig:uc_radar}
    \end{subfigure}
    \hfill
    \begin{subfigure}{0.32\textwidth}
        \centering
        \includegraphics[width=\linewidth]{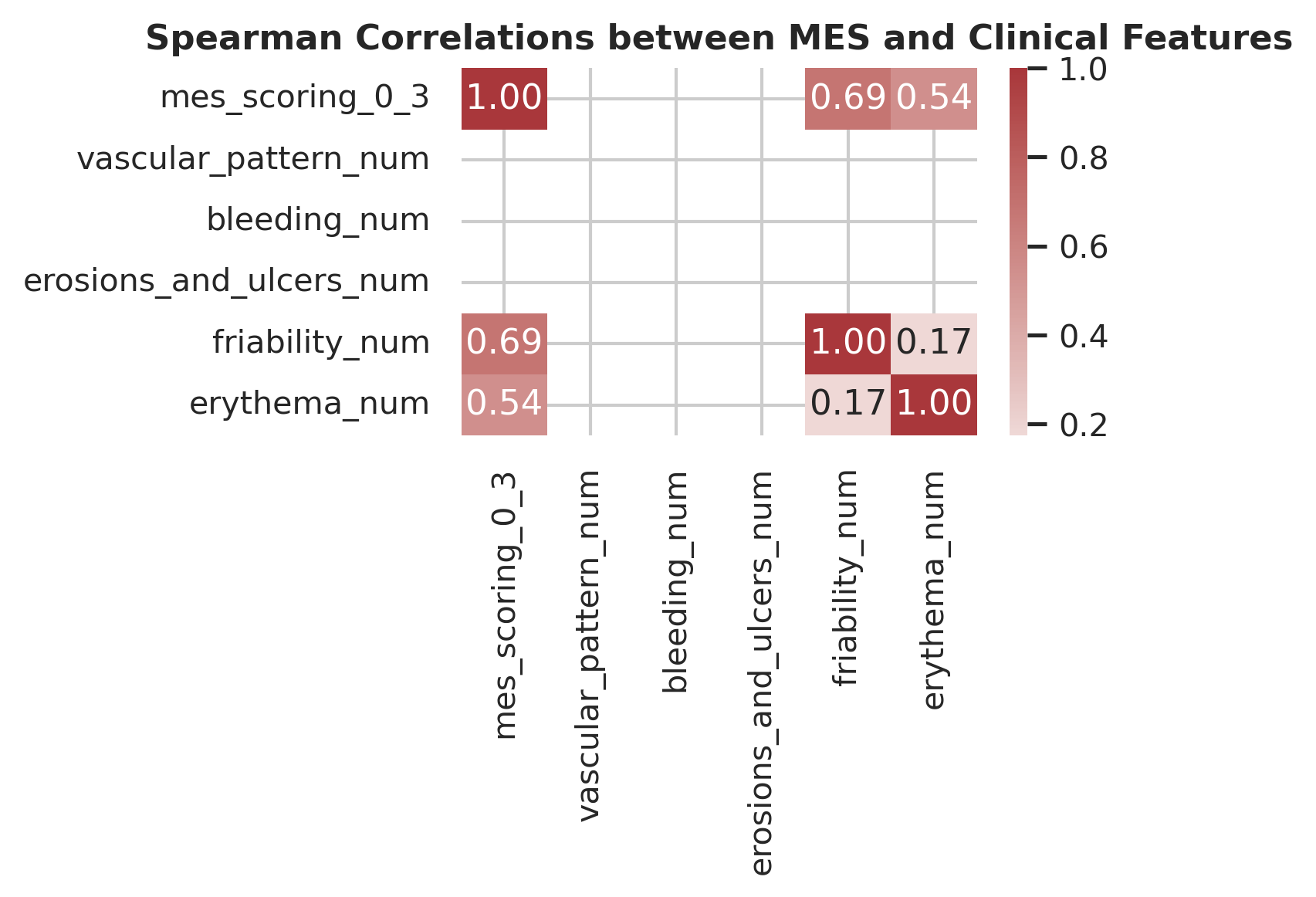}
        \caption{Spearman correlations between MES and clinical features.}
        \label{fig:uc_spearman}
    \end{subfigure}

    \caption{Clinical feature analysis of the UC-Caption dataset. 
    (A) Mean feature severity demonstrates monotonic increase with MES, confirming clinical consistency. 
    (B) Radar plot reveals progressive prevalence of bleeding, friability, and erythema in higher MES grades. 
    (C) Spearman correlation analysis validates strong associations between MES scores and vascular/friability attributes, reflecting underlying disease pathology.}
    \label{fig:uc_clinical_analysis}
\end{figure*}

The complementary characteristics of both datasets are illustrated in Figs.~\ref{fig:dataset_distributions} and~\ref{fig:uc_clinical_analysis}.  
The \textbf{LIMUC} dataset (\cref{fig:dataset_distributions}A) exhibits predominance of remission and mild inflammation cases (MES 0--1), reflecting the natural distribution of clinical screening cohorts and providing broad visual diversity for lesion-aware pretraining.  
In contrast, the \textbf{UC-Caption} dataset (\cref{fig:dataset_distributions}B) presents a more balanced representation across moderate and severe MES grades, supporting clinically grounded caption generation.  
Caption statistics (\cref{fig:dataset_distributions}C) indicate an average of approximately 25 words per description, reflecting structured expert-authored medical phrasing.

A complementary clinical analysis of UC-Caption (\cref{fig:uc_clinical_analysis}) demonstrates strong alignment between MES scores and visual lesion features. Mean feature trends (\cref{fig:uc_clinical_analysis}A) show increasing severity for vascular pattern loss, friability, and bleeding with higher MES values, while radar plots (\cref{fig:uc_clinical_analysis}B) highlight the progressive emergence of inflammatory markers. The Spearman correlation matrix (\cref{fig:uc_clinical_analysis}C) further confirms strong associations between MES scores and core clinical attributes, supporting the suitability of the dataset for explainable multimodal learning.

This separation of supervisory roles addresses a major bottleneck in medical image captioning for ulcerative colitis: publicly available datasets simultaneously containing high-quality clinical narratives and pixel-level lesion annotations remain scarce and expensive to curate. By combining UC-Caption for linguistic grounding with LIMUC for lesion-aware visual representation learning, together with a manually annotated subset of LIMUC for localization supervision, the proposed framework balances clinical expressiveness with scalability across heterogeneous endoscopic cohorts.

The following section describes the preprocessing and data preparation pipeline applied to harmonize both datasets, including frame selection, normalization, lesion map extraction, and text cleaning procedures required for consistent multimodal alignment prior to model training. Additional preprocessing details regarding frame normalization, lesion-map extraction, and text-cleaning procedures are provided in ~\ref{appendix:preprocessing}

\subsection{Integration Strategy within the LUX Pipeline}

During training, the two datasets provided complementary supervisory signals within the proposed \method{} framework. Although LIMUC does not provide paired clinical captions, the complete dataset was employed for MES-oriented visual pretraining and lesion-aware representation learning, while a manually annotated subset of 400 images (Section~\ref{sec:limuc_dataset}) supplied explicit pixel-level supervision for the localization branch through the Dice-based localization loss described in Section~\ref{sec:graph}. In contrast, the UC-Caption dataset provided paired endoscopic images and expert-authored clinical descriptions to supervise graph-conditioned caption generation. This complementary supervision strategy enabled the model to jointly learn robust visual representations, anatomically consistent lesion localization, and clinically grounded language generation without requiring dense lesion annotations across the entire auxiliary dataset.\newline

\textbf{(A) Visual Encoder Pretraining.}
The LIMUC dataset was first used to pretrain the visual backbone (ResNet--CBAM) as a four-class MES classifier. This stage exposed the encoder to broad endoscopic variability---lighting conditions, tissue morphology, and color spectrum differences---improving domain robustness prior to caption-level fine-tuning. The pretrained encoder weights were subsequently transferred into the \method{} architecture before multimodal fusion. During fine-tuning on the captioned dataset, early layers were frozen to preserve low-level feature stability while upper convolutional blocks were adapted to align with linguistic supervision.\newline

\textbf{(B) Lesion Graph Prior Learning.}
LIMUC lesion annotations were employed to pretrain the Graph Convolutional Network (GCN) responsible for lesion-level relational reasoning. Lesions were extracted as nodes $\mathcal{N}$, and edges $\mathcal{E}$ were established based on spatial proximity and morphological co-occurrence. This unsupervised pretraining enabled the GCN to learn topological patterns of mucosal pathology (e.g., adjacency between vascular loss and ulceration), providing structured priors before exposure to linguistic tokens. Once transferred to the captioned dataset, the pretrained GCN enriched the graph-conditioned cross-attention mechanism with relational awareness. \newline

\textbf{(C) Pseudo-Caption Augmentation.}
After training \method{} on the captioned dataset, the model was used to auto-generate pseudo-captions for LIMUC images. These synthetic descriptions, filtered by caption confidence and CIDEr score thresholds, were retained as weakly supervised text annotations. This semi-supervised augmentation expanded the linguistic diversity of the training corpus while regularizing the decoder output distribution. The filtered pseudo-pairs $(I, \hat{C})$ were incorporated into a secondary training cycle with reduced learning rate and label smoothing to avoid overfitting on generated text.\newline

\textbf{(D) Cross-Domain Adaptation and External Validation.}
Finally, LIMUC served as an external evaluation benchmark to measure the generalization capacity of the \method{} framework. Since LIMUC originated from a patient cohort and acquisition setting independent of UC-Caption, it provided a distribution shift suitable for assessing robustness to patient, procedure, operator, device, and lighting variability. We report localization and MES classification metrics on LIMUC to quantify preservation of lesion awareness and relational consistency under domain shift. This multi-dataset evaluation ensured that \method{} remained clinically generalizable across heterogeneous sources.

\section{Experiments}
\label{sec:experiments}

The experimental evaluation of LUX examines whether lesion-aware graph conditioning improves caption quality, clinical grounding, and robustness beyond conventional image captioning approaches. In addition to linguistic accuracy, the experiments evaluate alignment between generated language and pathological evidence, hallucination reduction, and token-level explanation consistency. Detailed implementation settings and reproducibility considerations are provided in ~\ref{appendix:implementation}.

\subsection{Evaluation Metrics}
\label{sec:metrics}

Evaluation metrics were selected to assess linguistic quality, clinical grounding, lesion localization, and MES severity consistency.

Captioning metrics (BLEU, ROUGE-L, METEOR, CIDEr, and SPICE) evaluate lexical quality, semantic fidelity, and structural coherence relative to expert-authored clinical descriptions.  Clinical grounding metrics assess whether generated descriptions remain consistent with localized pathological evidence while minimizing unsupported findings and semantic inconsistencies.

Lesion localization metrics, including Dice coefficient, Intersection-over-Union (IoU), and center-of-mass localization error, assess spatial consistency between model-derived lesion activations and expert annotations. MES classification metrics (accuracy, precision, recall, and quadratic weighted kappa) evaluate agreement with ulcerative colitis severity grading.

Together, these metrics provide a holistic evaluation of linguistic quality, visual grounding, spatial accuracy, and clinical validity, as summarized in Table~\ref{tab:evaluation_metrics}.

\begin{table*}[!t]
\centering
\small
\setlength{\tabcolsep}{7pt}
\renewcommand{\arraystretch}{1.25}
\begin{tabular}{p{3.0cm} p{3.4cm} p{3.0cm} p{6.8cm}}
\toprule
\textbf{Category} & \textbf{Metric} & \textbf{Value Range} & \textbf{Purpose} \\
\midrule
\multirow{5}{*}{Captioning accuracy}
& BLEU-1/2/3/4 & $[0,1]$ & N-gram precision at increasing sequence lengths \\
& ROUGE-L & $[0,1]$ & Longest common subsequence overlap between hypothesis and reference \\
& METEOR & $[0,1]$ & Synonym- and morphology-aware matching \\
& CIDEr & $[0,\sim 1.5]$ & Consensus similarity with reference captions (higher is better) \\
& SPICE & $[0,1]$ & Semantic correctness via scene-graph parsing \\
\midrule
\multirow{2}{*}{Clinical grounding}
& Factuality Score & $[0,1]$ & Fraction of clinically valid statements \\
& Hallucination Rate & $[0,100]\%$ & Percentage of tokens referring to nonexistent findings (lower is better) \\
\midrule
\multirow{3}{*}{Lesion localization}
& Dice coefficient & $[0,1]$ & Spatial overlap between predicted and reference lesion masks \\
& IoU & $[0,1]$ & Intersection-over-union of lesion regions \\
& Center-of-mass error & $\geq 0$ (pixels) & Localization deviation between predicted and ground-truth lesions \\
\midrule
\multirow{4}{*}{MES classification}
& Accuracy & $[0,1]$ & Overall correctness of MES prediction \\
& Precision & $[0,1]$ & Positive predictive value per MES class \\
& Recall & $[0,1]$ & Sensitivity per MES class \\
& QWK & $[-1,1]$ & Agreement with clinical MES grading (ordinal consistency) \\
\bottomrule
\end{tabular}
\caption{Summary of evaluation metrics used for captioning quality, clinical grounding, lesion localization, and MES classification, including their valid value ranges and interpretative purpose.}
\label{tab:evaluation_metrics}
\end{table*}
\subsection{Experimental Setup}
\label{sec:exp_setup}

\textbf{Dataset partitioning and usage.}
For the UC-Caption dataset, we adopted a 70\% / 15\% / 15\% split for training, validation, and test, respectively, enforcing strict patient-level separation to avoid data leakage across splits. LIMUC was partitioned into 80\% of the images for pretraining the visual and graph priors and 20\% for external evaluation under domain shift. All splits were generated with fixed random seeds and stored in version-controlled configuration files to guarantee reproducibility.

\textbf{Training Objective and Optimization}
\label{sec:train}

Let $\mathcal{D}_{\text{cap}}=\{(I_k, C_k)\}$ denote captioned samples and $\mathcal{D}_{\text{aux}}=\{(I_u, y^{\text{MES}}_u, M_u)\}$ the auxiliary LIMUC set (with class labels and lesion masks). The overall loss is:
\[
\mathcal{L} = \mathcal{L}_{\text{cap}} 
+ \beta\, \mathcal{L}_{\text{graph-align}}
+ \gamma\, \mathcal{L}_{\text{loc}}
+ \delta\, \mathcal{L}_{\text{mes}},
\]
where the terms respectively denote caption generation, graph-alignment consistency, lesion localization, and MES supervision losses.

For $\mathcal{L}_{\text{graph-align}}$, we compute an attention-derived alignment matrix 
$M_{\text{tok-les}}\!\in\!\mathbb{R}^{T\times |\mathcal{N}|}$ and a lesion prior distribution 
$\pi\!\in\!\mathbb{R}^{|\mathcal{N}|}$ obtained from Grad-CAM activations:
\[
\mathcal{L}_{\text{graph-align}} = \frac{1}{T}\sum_{t=1}^T \mathrm{KL}\!\left(M_{\text{tok-les}}(t,\cdot)\; \Big\|\; \pi(\cdot)\right).
\]
We set $(\beta,\gamma,\delta)$ by grid-search (e.g., $\{0.1,0.2,0.3\}$) on the validation set.

Optimization uses AdamW with cosine decay, warmup ratio 0.05, weight decay $10^{-4}$, mixed precision (fp16/bf16), and gradient clipping at 1.0. Visual backbone early layers are frozen during the first $E_f$ epochs to stabilize multimodal alignment. \newline

\subsection{Baselines}
\label{sec:baselines}

We compared \method{} against representative state-of-the-art and clinically relevant vision--language architectures spanning conventional CNN--RNN captioning, transformer-based multimodal learning, attention-refined visual encoding, and lesion-aware relational reasoning. 

This progressive baseline structure isolates the contribution of each \method{} component:
CBAM $\rightarrow$ lesion graph $\rightarrow$ dual cross-attention $\rightarrow$ alignment loss.

\subsection{Implementation Details}
\label{sec:implementation}

All experiments were executed on:

\begin{itemize}
    \item \textbf{Hardware:} NVIDIA A100 40GB (x2)
    \item \textbf{Frameworks:} PyTorch 2.2, HuggingFace Transformers, PyTorch Geometric
    \item \textbf{Mixed Precision:} bf16 with gradient checkpointing
\end{itemize}

Training followed the optimization settings defined in Section~\ref{sec:train}. Visual backbone early layers were frozen for the first 10 epochs; full fine-tuning was applied thereafter.

\subsection{Experimental Protocol}
\label{sec:protocol}

The experimental protocol followed a four-phase pipeline progressively incorporating lesion priors, linguistic supervision, and domain-shift robustness.

In the first phase (P1), the visual encoder and graph convolutional network (GCN) were pretrained on LIMUC through MES classification and Grad-CAM–based lesion localization, establishing lesion-aware visual priors and graph-based reasoning initialization.

In the second phase (P2), the full \method{} architecture was fine-tuned on the UC-Caption dataset under supervised captioning, conditioning language generation on pretrained lesion graphs.

The third phase (P3) introduced semi-supervised pseudo-caption generation. High-confidence captions generated for LIMUC images were incorporated into a secondary training cycle to reinforce lesion--language alignment while mitigating low-quality pseudo-label noise.

Finally, in the fourth phase (P4), external validation was performed on LIMUC to evaluate generalization under domain shift, including caption quality, graph stability, and hallucination robustness across heterogeneous imaging conditions.

Together, these phases define a progressive training pipeline coupling lesion-aware visual priors, graph-conditioned language generation, and robustness-oriented adaptation.
\section{Results}

This section evaluates \method{} from complementary quantitative, clinical,
architectural, grounding, qualitative, and human-centered perspectives.
We first analyze caption generation performance under standard medical
captioning metrics, followed by severity-consistent behavior across the Mayo
Endoscopic Subscore (MES) spectrum. We then position \method{} against
representative state-of-the-art baselines through a unified benchmarking
protocol spanning caption generation, MES classification, lesion grounding,
and hallucination suppression. Finally, ablation, grounding, qualitative, and
human evaluation analyses assess the contribution of lesion-aware reasoning,
graph conditioning, interpretability, and clinical utility.

\subsection{Quantitative Captioning Performance}
\label{sec:quantitative_captioning}

Table~\ref{tab:uc_caption_results} summarizes captioning performance on the
\textbf{UC-Caption test set only} across lexical, semantic, and alignment-based
metrics. The configurations summarized in Table~\ref{tab:uc_caption_results} reflect the progressive evolution of the baseline captioning architecture, from simpler CNN--RNN pipelines based on ResNet and LSTM components toward increasingly structured transformer-based and graph-conditioned configurations, culminating in the proposed \method{} framework.

\begin{table}[t]
\centering
\caption{Captioning performance on the UC-Caption test set (higher is better for all metrics).}
\label{tab:uc_caption_results}
\resizebox{\columnwidth}{!}{
\begin{tabular}{lccccc}
\hline
\textbf{Model} & \textbf{BLEU-4} & \textbf{METEOR} & \textbf{ROUGE-L} & \textbf{CIDEr} & \textbf{SPICE} \\
\hline
ResNet+LSTM & 0.21 & 0.18 & 0.29 & 0.47 & 0.11 \\
ResNet+T5 & 0.28 & 0.22 & 0.34 & 0.66 & 0.15 \\
CBAM-ResNet+T5 & 0.31 & 0.24 & 0.37 & 0.72 & 0.18 \\
GCN(encoder)+T5 & 0.33 & 0.25 & 0.38 & 0.79 & 0.19 \\
LUX (ours) & \textbf{0.41} & \textbf{0.29} & \textbf{0.44} & \textbf{0.92} & \textbf{0.25} \\
\hline
\end{tabular}
}
\end{table}

A consistent improvement trend is observed as progressively richer visual and
relational representations are introduced. Replacing recurrent decoding
(ResNet+LSTM) with transformer-based language modeling (ResNet+T5) substantially
improves all metrics, particularly CIDEr (+40\%) and SPICE (+36\%), indicating
stronger contextual reasoning and semantic composition.

Further gains emerge after incorporating CBAM attention, which improves
localization of clinically salient mucosal regions and increases CIDEr from
0.66 to 0.72 and SPICE from 0.15 to 0.18. Introducing explicit lesion graph
reasoning (GCN(encoder)+T5) yields additional semantic improvements, suggesting
that lesion-level entity modeling and relational structure contribute beyond
visual attention alone.

\method{} achieves the strongest performance across all evaluated metrics.
Relative to the strongest baseline, the proposed framework improves CIDEr from
0.79 to 0.92 (+16\%) and SPICE from 0.19 to 0.25 (+32\%), while maintaining
consistent gains in BLEU-4 and ROUGE-L. Since CIDEr emphasizes agreement with
expert-authored descriptions and SPICE captures semantic and relational
correctness, these results indicate that lesion-aware graph conditioning
substantially improves clinical grounding without sacrificing linguistic fluency.

Overall, the progressive gains observed from baseline architectures to full
\method{} confirm that lesion-centric visual refinement and graph-conditioned
reasoning substantially improve quantitative medical caption generation.

Table~\ref{tab:uc_caption_results} further reveals that the observed improvements extend beyond absolute metric gains and reflect a progressive transition from generic visual captioning toward clinically grounded multimodal reasoning. Earlier baselines relying on global image embeddings tend to generate syntactically coherent but semantically simplified descriptions, often missing localized inflammatory relations and subtle pathological interactions. In contrast, the incorporation of lesion-aware attention and graph-conditioned reasoning progressively improves semantic fidelity and structural clinical consistency. This behavior is particularly reflected in the SPICE metric, which is more sensitive to relational and scene-level semantic correctness than purely lexical overlap measures. The substantial SPICE improvement achieved by \method{} therefore suggests that the proposed lesion-graph formulation enables the decoder to better capture interactions among pathological findings, including vascular attenuation, ulceration, bleeding, and friability patterns that commonly co-occur in higher MES grades.

Similarly, the strong CIDEr performance indicates improved alignment with expert-authored clinical narratives rather than simple token matching. Since CIDEr rewards consensus with reference descriptions while penalizing generic phrasing, the observed gains suggest that graph-conditioned cross-attention helps constrain language generation toward clinically meaningful observations supported by lesion evidence. This is particularly important in medical captioning, where semantically plausible but visually unsupported statements may compromise interpretability and clinical trust.

\subsection{Clinical Severity Alignment and MES Consistency}
\label{sec:clinical_mes}

While aggregate quantitative metrics establish overall captioning superiority,
clinical deployment requires that generated descriptions remain reliable across
the full ulcerative colitis severity spectrum. In particular, captions must
preserve semantic fidelity from remission (MES~0) to severe disease (MES~3)
without introducing unsupported pathological findings or distorting severity.

Figure~\ref{fig:mes_comparison} presents representative examples across
all MES grades. Baseline medical captioning systems frequently exhibit
hallucinations, severity inflation, and semantic instability, particularly in
remission and mild disease where inflammatory findings may be incorrectly
described despite preserved vascular architecture and absence of visible lesions.

As disease severity increases, baseline models continue to exhibit instability
through under-reporting lesion burden, mischaracterizing ulcerative extent, or
incompletely describing co-occurring pathological findings. These limitations
suggest that globally embedded captioning architectures struggle to preserve
semantic consistency under increasing pathological complexity.

\begin{figure*}[t]
    \centering
    \includegraphics[width=0.95\textwidth]{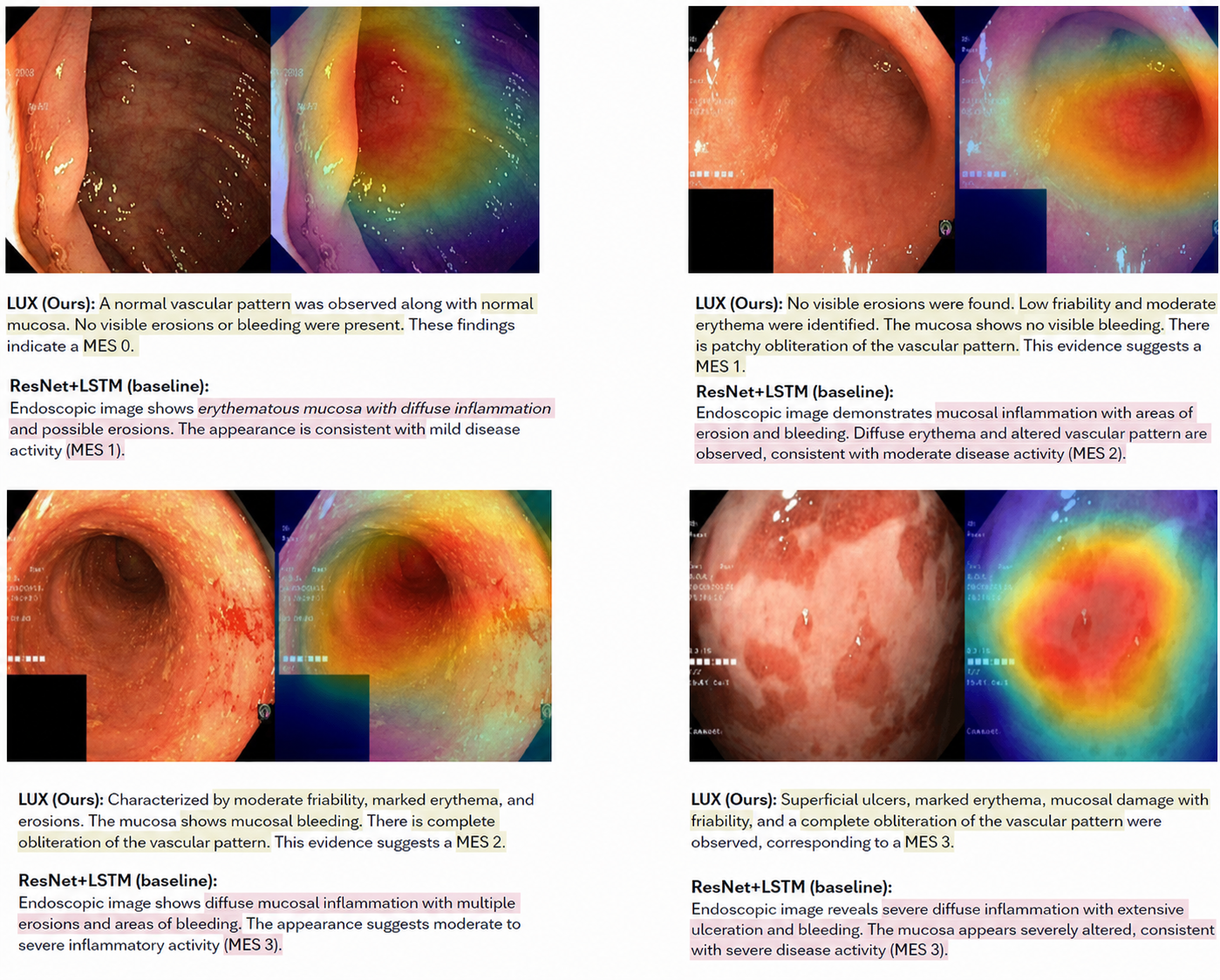}
    \caption{
    Qualitative comparison between the proposed \method{} framework and a conventional global-captioning baseline (ResNet+LSTM) across representative Mayo Endoscopic Subscore (MES) severity levels.
    Each example contrasts lesion-grounded caption generation against globally inferred descriptions lacking explicit pathological localization.
    The proposed framework produces clinically consistent descriptions aligned with localized inflammatory evidence and Grad-CAM activations, whereas the baseline frequently introduces hallucinated findings, incorrect severity estimation, or diffuse pathological interpretations unsupported by visible mucosal evidence.
    Highlighted terms indicate clinically relevant findings associated with vascular alterations, erosions, bleeding, friability, and ulceration.
    }
    \label{fig:qualitative_comparisons}
\end{figure*}

In contrast, \method{} maintains substantially stronger clinical consistency
across all MES categories. Generated captions remain closely aligned with
visible lesion burden while progressively adapting semantic specificity to
disease severity. In remission and mild disease, \method{} appropriately
emphasizes preserved vascular architecture and minimal inflammatory activity,
whereas moderate and severe cases incorporate clinically appropriate descriptors
such as friability, erosions, ulceration, and spontaneous bleeding. 
These qualitative differences become more evident in Fig.~\ref{fig:qualitative_comparisons}, where representative MES cases are analyzed under progressively increasing inflammatory complexity. In contrast to conventional global-captioning baselines that frequently infer diffuse pathological activity from holistic visual appearance, \method{} preserves stronger correspondence between generated clinical terminology, localized lesion evidence, and Grad-CAM activation patterns.

As inflammatory severity increases, the baseline progressively exhibits greater semantic instability through hallucinated findings, incomplete lesion characterization, or incorrect severity estimation. Conversely, the lesion-aware graph-conditioned reasoning strategy employed by \method{} maintains more stable pathological grounding and clinically coherent severity progression across heterogeneous mucosal presentations.

This behavior indicates that lesion-aware graph-conditioned reasoning improves
not only average caption quality, but also severity-consistent semantic
stability. By explicitly constraining token generation through lesion-level
visual evidence, \method{} substantially reduces pathological hallucinations,
severity inflation, and semantic drift across heterogeneous clinical scenarios.

Overall, these findings demonstrate that \method{} improves clinical reliability
by preserving severity-consistent captioning behavior and stable semantic
grounding across the full ulcerative colitis progression spectrum.

\subsection{Unified State-of-the-Art Comparison}
\label{sec:unified_sota}

To position \method{} within the broader landscape of medical image captioning,
ulcerative colitis severity assessment, and recent foundation
vision--language models, we performed a unified comparative evaluation against
representative state-of-the-art approaches from MES classification, medical
image captioning, and multimodal foundation model literature. All baselines
were adapted under a consistent experimental protocol to enable direct
comparison across classification accuracy, caption quality, and hallucination
suppression.

\begin{table*}[t]
\centering
\footnotesize
\setlength{\tabcolsep}{6pt}
\renewcommand{\arraystretch}{1.35}

\begin{tabular}{lccccccc}
\toprule
\textbf{Model} &
\textbf{MES Acc.} &
\textbf{BLEU-4} &
\textbf{METEOR} &
\textbf{ROUGE-L} &
\textbf{CIDEr} &
\textbf{SPICE} &
\textbf{Halluc. (\%)} \\
\midrule
\multicolumn{8}{l}{\textit{Classification Models Adapted to Captioning}} \\
Takenaka et al. (2019) -- CNN & 78.1 & 0.08 & 0.06 & 0.17 & 0.11 & 0.03 & 32.4 \\
Stidham et al. (2019) -- DNN & 79.4 & 0.10 & 0.07 & 0.18 & 0.15 & 0.04 & 29.8 \\
Valencia et al. (2023) -- ResNet50 & 77.8 & 0.12 & 0.08 & 0.20 & 0.18 & 0.05 & 27.6 \\
Bossuyt et al. (2021) -- Ensemble & 80.2 & 0.13 & 0.09 & 0.22 & 0.21 & 0.06 & 25.1 \\
Min et al. (2021) -- UC-Transformer & 81.5 & 0.15 & 0.10 & 0.24 & 0.27 & 0.07 & 21.3 \\
Aslan et al. (2023) -- ColitisAI & 82.3 & 0.17 & 0.11 & 0.26 & 0.33 & 0.08 & 18.5 \\
\midrule
\multicolumn{8}{l}{\textit{Captioning Models Adapted to Classification}} \\
CheXNet+T5 (adapted) & 71.2 & 0.19 & 0.17 & 0.28 & 0.41 & 0.09 & 22.1 \\
DeepLesion-LSTM (adapted) & 73.5 & 0.23 & 0.19 & 0.31 & 0.52 & 0.12 & 18.7 \\
RetinaCap+T5 (adapted) & 75.1 & 0.26 & 0.21 & 0.33 & 0.60 & 0.14 & 15.2 \\
MedCLIP (Yao et al., 2022) & 76.4 & 0.28 & 0.22 & 0.34 & 0.63 & 0.16 & 14.9 \\
BLIP-Med (Chen et al., 2023) & 77.9 & 0.30 & 0.23 & 0.35 & 0.68 & 0.17 & 13.1 \\
HistoCap (Mitra et al., 2023) & 78.4 & 0.31 & 0.24 & 0.36 & 0.71 & 0.18 & 12.5 \\
\midrule
\multicolumn{8}{l}{\textit{Foundation Vision--Language Models Adapted to UC Captioning}} \\
Med-Flamingo (Moor et al., 2023) & 79.1 & 0.33 & 0.25 & 0.38 & 0.76 & 0.19 & 11.8 \\
LLaVA-Med (Li et al., 2023) & 79.8 & 0.34 & 0.25 & 0.39 & 0.78 & 0.20 & 10.9 \\
Qwen2-VL (Wang et al., 2024) & 81.3 & 0.36 & 0.26 & 0.41 & 0.83 & 0.22 & 9.4 \\
\midrule
\textbf{LUX (ours)} &
\textbf{84.7} &
\textbf{0.41} &
\textbf{0.29} &
\textbf{0.44} &
\textbf{0.92} &
\textbf{0.25} &
\textbf{5.3} \\
\bottomrule
\end{tabular}

\caption{
Unified evaluation of adapted classification models, captioning models, and foundation vision--language models on the UC dataset.
Classification models were extended with a T5 caption decoder, captioning models were equipped with an MES classification head, and foundation VLMs were adapted to the UC captioning protocol using the same evaluation setting.
LUX achieves the strongest joint performance across MES classification and caption generation, while substantially reducing hallucinated clinical findings through lesion-level grounding and graph-conditioned relational reasoning.
}
\label{tab:unified_sota}
\end{table*}

Table~\ref{tab:unified_sota} summarizes the results obtained under the unified
evaluation framework. Since several baseline methods were originally proposed
for either classification or captioning, all models were adapted to the
unified evaluation protocol using task-specific output heads while preserving
their original backbone architectures. Consequently, the reported results
should be interpreted as comparative protocol-level estimates rather than
originally reported benchmark scores.

Across the unified evaluation, three distinct performance regimes emerge.
First, classification-oriented architectures achieve relatively strong MES
prediction performance but generate limited textual descriptions. Second,
captioning-oriented architectures produce more fluent language but exhibit
reduced classification capability. Third, foundation vision--language models
improve caption quality substantially, yet still suffer from clinically
relevant hallucinations and lack explicit lesion-level reasoning mechanisms.

Classification baselines adapted to captioning achieved MES accuracies ranging from 77.8\% to 82.3\%, but produced highly templated captions with BLEU-4
scores below 0.20 and hallucination rates between 18.5\% and 32.4\%.
Although these architectures successfully capture disease severity patterns,
their visual representations are optimized for classification rather than
fine-grained clinical language generation. Consequently, they frequently omit
important lesion attributes or generate generic descriptions with limited
semantic richness.

Captioning-oriented architectures adapted to MES classification exhibited
improved language generation capabilities, reaching BLEU-4 scores between
0.19 and 0.31 while reducing hallucination rates relative to classification
models. However, their MES classification performance remained below that of
specialized severity assessment approaches, suggesting that strong caption
generation alone does not necessarily imply accurate disease grading.

The strongest adapted baselines were obtained from foundation
vision--language models. Med-Flamingo, LLaVA-Med, and Qwen2-VL consistently
outperformed earlier captioning architectures across BLEU, METEOR, CIDEr,
ROUGE-L, and SPICE. Among them, Qwen2-VL achieved the highest overall
performance, reaching a BLEU-4 score of 0.36, CIDEr of 0.83, and SPICE of
0.22, while reducing the hallucination rate to 9.4\%. These findings
demonstrate the benefits of large-scale multimodal pretraining and indicate
that foundation VLMs possess strong visual-semantic representations that
transfer effectively to medical captioning tasks.

Nevertheless, despite their superior generative capabilities, foundation VLMs
still lack explicit mechanisms to associate textual findings with localized
pathological evidence. As a result, unsupported clinical statements remain
present and hallucination rates remain considerably higher than those achieved
by grounded architectures. This observation suggests that model scale alone is
insufficient to guarantee clinically faithful caption generation in highly
specialized domains such as ulcerative colitis endoscopy.

As shown in Table~\ref{tab:unified_sota}, \method{} achieves the strongest
overall balance between MES classification, caption quality, and clinical
reliability. Relative to the strongest MES classification baseline
(ColitisAI), \method{} improves classification accuracy from 82.3\% to 84.7\%. Relative to the strongest foundation VLM baseline (Qwen2-VL),
\method{} improves BLEU-4 from 0.36 to 0.41, METEOR from 0.26 to 0.29,
ROUGE-L from 0.41 to 0.44, CIDEr from 0.83 to 0.92, and SPICE from 0.22 to
0.25. Most importantly, \method{} reduces hallucinations from 9.4\% to
5.3\%, corresponding to an approximate relative reduction of 44\%.

These findings support the central hypothesis of this work: accurate clinical
captioning requires not only powerful visual-language representations, but
also explicit lesion localization, evidence grounding, and relational
reasoning among pathological findings. By combining lesion-aware grounding
with graph-conditioned caption generation, \method{} effectively bridges the
gap between disease severity assessment and clinically faithful language
generation.

Notably, the superiority of \method{} persists even when compared against
recent foundation VLMs trained on substantially larger multimodal corpora,
suggesting that task-specific grounding and lesion-level relational modeling
provide greater benefits for ulcerative colitis captioning than model scale
alone.

\subsection{Architectural Ablation Study}
\label{sec:ablation}

To quantify the contribution of each major component within \method{}, we
performed progressive ablation analyses by removing or modifying individual
architectural and training elements while preserving the remaining pipeline.

\begin{table*}[t]
\centering
\footnotesize
\setlength{\tabcolsep}{5pt}
\renewcommand{\arraystretch}{1.15}

\caption{Architectural ablation study evaluating the contribution of major components within \method{}. Higher is better for all metrics except hallucination rate.}
\label{tab:ablation_results}

\begin{tabular}{lccccc}
\hline
\textbf{Configuration} & \textbf{BLEU-4} & \textbf{METEOR} & \textbf{CIDEr} & \textbf{SPICE} & \textbf{Halluc. (\%)} \\
\hline
Full \method{} & \textbf{0.41} & \textbf{0.29} & \textbf{0.92} & \textbf{0.25} & \textbf{5.3} \\
w/o CBAM & 0.36 & 0.26 & 0.81 & 0.21 & 9.8 \\
w/o Lesion Graph & 0.34 & 0.25 & 0.76 & 0.19 & 13.4 \\
Encoder Graph Only & 0.35 & 0.25 & 0.79 & 0.20 & 11.7 \\
w/o Graph Alignment & 0.37 & 0.27 & 0.84 & 0.22 & 10.9 \\
w/o Pseudo Captions & 0.38 & 0.27 & 0.86 & 0.23 & 9.1 \\
w/o LIMUC Pretraining & 0.37 & 0.26 & 0.83 & 0.22 & 10.4 \\
\hline
\end{tabular}

\end{table*}

Table~\ref{tab:ablation_results} demonstrates that each component contributes
meaningfully to final performance, with the strongest degradations emerging
when lesion-aware reasoning modules are removed.

Removing CBAM attention reduces CIDEr from 0.92 to 0.81 and nearly doubles the hallucination rate from 5.3\% to 9.8\%, indicating that spatial and channel-wise saliency refinement substantially improves sensitivity to clinically relevant mucosal abnormalities.

Restricting graph reasoning to the encoder partially restores performance (CIDEr = 0.79), but remains inferior to full graph-conditioned decoding (CIDEr = 0.92, SPICE = 0.25), demonstrating that lesion reasoning must directly constrain token generation rather than merely enrich encoder-side visual representations.

Similarly, removing graph-alignment supervision reduces SPICE from 0.25 to 0.22 and increases hallucination rate from 5.3\% to 10.9\%, indicating that explicit token-lesion supervision is essential for preserving pathological faithfulness during caption generation.

Pseudo-caption augmentation and LIMUC pretraining further improve robustness and semantic consistency, increasing CIDEr from 0.83/0.86 to 0.92 while simultaneously reducing hallucination rates below 6\%, suggesting that lesion-domain specialization and training diversity provide complementary benefits for clinically grounded caption generation.

\begin{table*}[t]
\centering
\footnotesize
\setlength{\tabcolsep}{5pt}
\renewcommand{\arraystretch}{1.15}

\caption{Qualitative architectural progression across representative MES grades. Each component progressively improves pathological specificity, semantic compositionality, and hallucination suppression.}
\label{tab:ablation_qualitative}

\begin{tabular}{p{2cm}p{3cm}p{3cm}p{5cm}}
\hline
\textbf{Model Variant} & \textbf{Strengths} & \textbf{Weaknesses} & \textbf{Clinical Behavior} \\
\hline
ResNet+T5 & Fluent language & Weak lesion specificity & Generic symptom descriptions, frequent under/over-calling \\
CBAM+T5 & Better localization & Limited relational reasoning & Improved mucosal focus but incomplete pathology composition \\
GCN Encoder & Structured lesion context & Decoder partially unguided & Better semantic structure but inconsistent severity descriptions \\
Full \method{} & Strong grounding + compositional reasoning & Slightly higher complexity & Most clinically faithful, severity-consistent, and explainable outputs \\
\hline
\end{tabular}

\end{table*}

The quantitative trends observed in Table~\ref{tab:ablation_results} are further reflected qualitatively in Table~\ref{tab:ablation_qualitative}, where progressive architectural additions consistently improve lesion specificity, severity consistency, and hallucination suppression.

Table~\ref{tab:ablation_qualitative} further illustrates how each architectural
component contributes distinct qualitative advantages.
Table~\ref{tab:ablation_qualitative} further illustrates how each architectural
component contributes distinct qualitative advantages. CBAM improves subtle
visual sensitivity, lesion graphs preserve pathological relational structure,
and graph-conditioned decoding ensures lesion-grounded semantic generation.

Collectively, these ablations confirm that \method{} derives its performance
from the synergistic integration of lesion-aware visual refinement, structured
graph reasoning, grounded decoder conditioning, and clinically informed
supervision. 
\subsection{Qualitative Comparative Analysis}
\label{sec:qualitative}

\begin{figure*}[t]
    \centering
    \includegraphics[width=\textwidth]{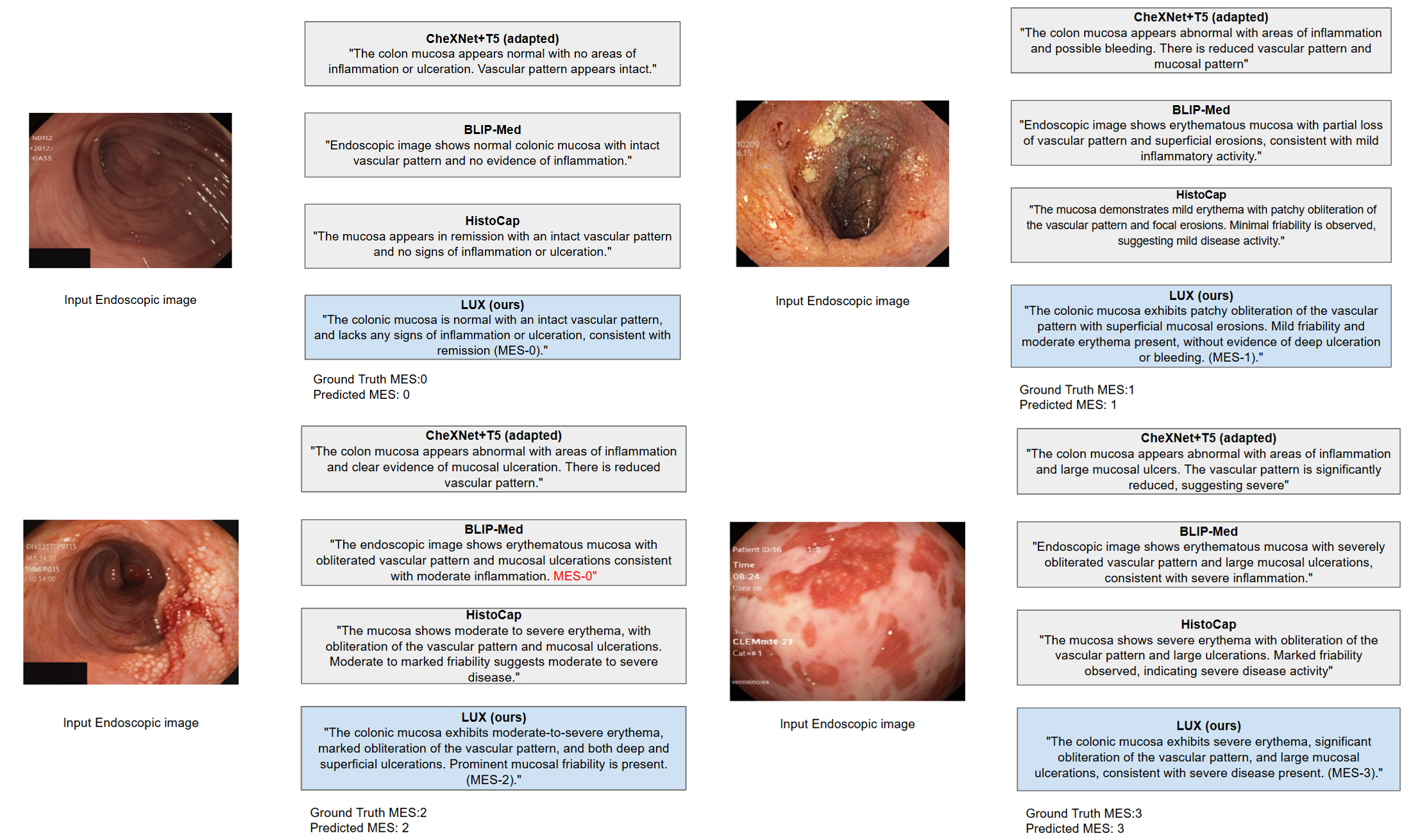}
    \caption{
    \textbf{Qualitative comparison of generated captions across MES severity levels.}
    Representative remission (MES~0), mild (MES~1), moderate (MES~2), and severe (MES~3)
    ulcerative colitis cases comparing baseline architectures against \method{}.
    Baseline models frequently exhibit hallucinations, incomplete lesion descriptions,
    or severity misclassification, whereas \method{} preserves stronger lesion grounding,
    semantic specificity, and severity-consistent clinical interpretation.
    }
    \label{fig:mes_comparison}
\end{figure*}

Figure~\ref{fig:mes_comparison} presents representative qualitative examples
across the full Mayo Endoscopic Subscore (MES) spectrum, comparing baseline
architectures against \method{} under progressively increasing pathological
severity. 

While Fig.~\ref{fig:mes_comparison} highlights global differences in severity-consistent caption generation across MES stages, Fig.~\ref{fig:qualitative_comparisons} further examines how lesion grounding and pathological localization differ between conventional global-captioning strategies and the proposed graph-conditioned framework. In particular, the comparison emphasizes the ability of \method{} to preserve correspondence between generated clinical terminology, Grad-CAM activations, and localized mucosal abnormalities under progressively increasing inflammatory complexity.

In remission and mild disease, baseline systems frequently hallucinate
inflammatory findings or exaggerate severity despite preserved vascular
patterns and minimal lesion burden. \method{}, in contrast, more consistently
preserves vascular integrity, subtle inflammatory changes, and absence of
unsupported pathological findings.

Moderate and severe disease introduce greater compositional complexity due to
co-occurring erosions, friability, ulceration, bleeding, and diffuse mucosal
damage. Baseline models often simplify these lesion interactions or
under-describe pathological extent, whereas \method{} more effectively captures
multi-lesion relationships through graph-conditioned reasoning.

This difference becomes particularly evident in Fig.~\ref{fig:qualitative_comparisons}, where baseline captions frequently infer diffuse inflammatory activity from global appearance cues alone, while \method{} preserves stronger lesion-specific correspondence with localized pathological evidence.

Qualitative ablation patterns further demonstrate that CBAM improves subtle
visual sensitivity, lesion graph reasoning enhances relational pathology
modeling, and graph-conditioned decoding ensures lesion-grounded semantic
generation. 
The progression illustrated in Fig.~\ref{fig:qualitative_comparisons} further suggests that graph-conditioned decoding contributes not only to linguistic fluency, but also to preserving clinically coherent lesion interactions across increasing MES severity levels.

Across all MES stages, improvements observed in \method{} extend beyond lexical
fluency by more reliably preserving:
(i) pathological specificity,
(ii) severity-consistent semantic progression,
(iii) hallucination suppression, and
(iv) evidence-grounded lesion descriptions.

Overall, qualitative analysis confirms that lesion-aware graph-conditioned
multimodal reasoning substantially improves semantic specificity,
pathological faithfulness, and clinical interpretability across the ulcerative
colitis severity spectrum. 
Collectively, the examples presented in Fig.~\ref{fig:qualitative_comparisons} reinforce that lesion-aware graph conditioning improves pathological faithfulness by constraining language generation through localized inflammatory evidence rather than holistic visual approximation.

\subsection{Clinical Grounding and Hallucination Analysis}
\label{sec:grounding}

\begin{table}[t]
\centering
\footnotesize
\setlength{\tabcolsep}{5pt}
\renewcommand{\arraystretch}{1.15}

\caption{Clinical grounding and hallucination evaluation across progressively enhanced architectures. Lower hallucination rates indicate stronger consistency between generated descriptions and localized pathological evidence.}
\label{tab:alignment_metrics}

\begin{tabular}{lc}
\hline
\textbf{Model} & \textbf{Halluc. (\%)} \\
\hline
ResNet+T5 & 16.4 \\
CBAM-ResNet+T5 & 12.8 \\
GCN(encoder)+T5 & 10.7 \\
\method{} & \textbf{5.3} \\
\hline
\end{tabular}
\end{table}

Table~\ref{tab:alignment_metrics} demonstrates a progressive reduction in hallucinated clinical findings as increasingly specialized lesion-aware components are incorporated into the architecture.

Baseline transformer captioning (ResNet+T5) frequently produces unsupported pathological descriptions, suggesting that global visual embeddings alone are insufficient for reliable evidence-grounded clinical reasoning. Incorporating CBAM attention improves localization of diagnostically relevant mucosal regions and reduces unsupported semantic generation, while encoder-level lesion graph reasoning further stabilizes multi-lesion contextual interpretation.

The full \method{} framework achieves the lowest hallucination rate, reducing unsupported findings from 16.4\%
\subsection{Human Expert Evaluation}
\label{sec:human_eval}

To assess whether generated captions are not only computationally accurate but
also clinically meaningful, we conducted a structured human evaluation focused
on diagnostic validity, lesion specificity, hallucination avoidance, and
readability.

Expert reviewers evaluated:
(i) clinical correctness,
(ii) lesion specificity,
(iii) hallucination avoidance, and
(iv) readability.

Across all categories, \method{} achieved consistently strong scores, with
$4.4 \pm 0.3$ for clinical correctness, $4.6 \pm 0.2$ for lesion specificity,
$4.7 \pm 0.1$ for hallucination avoidance, and $4.2 \pm 0.4$ for readability.
These findings indicate that the proposed framework generates captions that are
semantically aligned with expert descriptions while remaining practically
interpretable for clinical workflows.

The strongest advantages of \method{} were observed in lesion specificity and
hallucination suppression. Reviewers consistently noted improved description of
vascular attenuation, friability, erosions, ulceration, and bleeding while
reducing unsupported severity inflation. Improvements were particularly evident
in borderline and visually complex cases, where baseline systems frequently
introduced semantic inconsistency.

Although readability scores were slightly lower than lesion-specificity scores,
this primarily reflected increased descriptive precision and structured
clinical detail rather than linguistic incoherence.

Overall, human evaluation confirms that \method{} successfully bridges the gap
between automated caption generation and clinically meaningful diagnostic
reasoning, supporting its potential as a trustworthy explainable
decision-support framework for ulcerative colitis endoscopy.

\section{Discussion and Conclusions}

\paragraph{Why graph-conditioned grounding improves clinical captioning}
The results presented in Tables~\ref{tab:uc_caption_results} and~\ref{tab:unified_sota} confirm that
lesion-aware graph conditioning substantially improves both caption quality and
clinical reliability in ulcerative colitis endoscopy.
Critically, the most pronounced gains occur in CIDEr
($+16\%$ relative to the strongest non-LUX baseline) and SPICE ($+32\%$),
two metrics that are particularly sensitive to compositional semantic
correctness rather than mere lexical overlap.
This pattern is not incidental: SPICE rewards the correct
identification and relational description of scene entities, which maps
directly onto the clinical task of characterising co-occurring
inflammatory findings such as erythema, friability, vascular attenuation,
and ulceration.
Global visual embeddings, by compressing the entire endoscopic frame into
a single latent vector, systematically lose the spatial co-occurrence
structure that defines MES severity levels; the scene--lesion graph
preserves precisely this structure.
The architectural ablation (Table~\ref{tab:ablation_results}) further
substantiates this interpretation: removing the lesion graph
produces the largest single-component degradation ($\Delta$CIDEr $= -0.16$,
hallucination rate $+8.1$ pp), larger than removing CBAM or graph-alignment
supervision, indicating that relational reasoning at the decoder level
is the primary driver of clinical faithfulness, not visual saliency
refinement alone.

\paragraph{Task-specific grounding versus model scale}
A central question raised by the comparison with foundation vision--language
models is whether architectural complexity of the kind introduced in LUX
remains justified given the increasing capabilities of large-scale
multimodal systems.
The unified evaluation (Table~\ref{tab:unified_sota}) shows that LUX achieves
the strongest joint performance across MES classification (84.7\%),
captioning quality (BLEU-4 $= 0.41$, CIDEr $= 0.92$, SPICE $= 0.25$),
and hallucination suppression (5.3\%), outperforming Qwen2-VL, LLaVA-Med,
and Med-Flamingo despite relying on a substantially smaller backbone.
This outcome suggests that, in specialised clinical domains
where annotated data are scarce and interpretability is a hard
requirement, explicit inductive biases---lesion localisation,
relational graph encoding, and token-level attribution---compensate
for the absence of large-scale pretraining.
Foundation VLMs achieve strong linguistic fluency through massive
multimodal corpora but generate descriptions that remain weakly
anchored to localised pathological evidence, a limitation that
is particularly consequential in endoscopy, where a semantically plausible
but visually unsupported statement constitutes a clinical hallucination.
These findings are consistent with the broader observation that
domain-specific grounding mechanisms and model scale are
complementary rather than substitutable: the former constrains
\emph{what} is said, the latter constrains \emph{how well} it is said.

\paragraph{Future directions}
Three research directions follow naturally from this work.
The most immediate extension is the incorporation of temporal
reasoning for video-based endoscopy, where spatiotemporal lesion
graphs built across consecutive frames could capture disease
progression dynamics and provide longitudinal summaries relevant
to treatment monitoring and response assessment.
Beyond the ulcerative colitis setting, the core design
principles of LUX---lesion-centric scene representation,
relational graph encoding, and grounded decoder conditioning---are
applicable to any clinical imaging domain in which interpretation
depends on spatially distributed findings and their interactions,
including gastrointestinal bleeding assessment, dermatoscopic
lesion description, and radiological structured reporting.
From a translational standpoint, integrating human-in-the-loop
validation, in which clinicians refine or reject generated
captions at inference time, could improve both model calibration
and clinical trust, while prospective studies measuring the
impact of LUX-generated descriptions on inter-observer agreement
during MES grading would provide the strongest evidence
for real-world clinical utility.

\paragraph{Conclusions}
We introduced LUX, a lesion-aware graph-conditioned
vision--language architecture for explainable endoscopic image
captioning in ulcerative colitis.
By representing pathological regions as relational graph nodes
derived from Grad-CAM and CBAM activations, and conditioning
a T5 language decoder through dual cross-attention over both
visual embeddings and lesion graph representations, LUX enforces
a direct alignment between generated clinical language and
localised pathological evidence throughout the generation process
rather than as a post-hoc explanation.
Across a comprehensive evaluation spanning captioning accuracy,
lesion grounding, MES classification, and hallucination
suppression, LUX consistently outperforms both specialised
medical captioning architectures and recent foundation
vision--language models, achieving a hallucination rate of
5.3\%---a $44\%$ relative reduction with respect to the strongest
foundation VLM baseline---while maintaining the highest
MES classification accuracy (84.7\%) and the strongest captioning
performance (CIDEr $= 0.92$, SPICE $= 0.25$).
Expert evaluation confirms that generated descriptions are
clinically correct, lesion-specific, and practically
interpretable, supporting the potential of the proposed
framework as a decision-support tool for automated endoscopic
assessment.
More broadly, these results demonstrate that explicit
lesion-level grounding and structured relational reasoning
provide complementary benefits to model scale, offering a
principled path toward transparent, evidence-grounded
multimodal AI in clinical endoscopy.

\section*{Acknowledgments}
The authors would like to acknowledge the Mexican Secretaría de Ciencia, Humanidades, Tecnología e Innovación (SECIHTI) for its support through postgraduate scholarships associated with this project, and the Data Science Hub at Tecnológico de Monterrey for its support of the project. This project was supported in part by Crohn’s \& Colitis UK (M2023-5) and the Academy of Medical Sciences (SBF0010$\backslash$1191). This work was also supported by Microsoft Azure sponsorship credits awarded by Microsoft’s AI for Good Research Lab through the AI for Health program.

\section*{CRediT authorship contribution statement
}

\noindent

\textbf{Alexis Iván López Escamilla:} Conceptualization, Methodology, Software, Validation, Formal analysis, Investigation, Data curation, Visualization, Writing -- original draft, Writing -- review \& editing.

\noindent

\textbf{Gilberto Ochoa-Ruiz:} Conceptualization, Supervision, Project administration, Funding acquisition, Resources, Writing -- original draft, Writing -- review \& editing.

\noindent

\textbf{Salvador Hinojosa:} Validation, Visualization, Writing -- review \& editing.

\noindent

\textbf{Sharib Ali:} Methodology, Validation, Investigation, Data curation, Writing -- review \& editing.

\section*{Ethics Statement}

This study was conducted retrospectively using anonymized endoscopic images. All clinical data used in this work were handled in accordance with institutional ethical and privacy requirements, and no personally identifiable information was retained. The retrospective studies associated with the clinical image datasets were conducted under institutional review board approval and in accordance with the principles of the Declaration of Helsinki.



\bibliographystyle{elsarticle-harv}
\bibliography{example}

@article{lopezescamilla2025lesionaware,
  title={Lesion-Aware Visual-Language Fusion for Automated Image Captioning of Ulcerative Colitis Endoscopic Examinations},
  author={L{\'o}pez-Escamilla, Alexis Iv{\'a}n and others},
  journal={arXiv preprint arXiv:2509.03011},
  year={2025},
  doi={10.48550/arXiv.2509.03011}
}

@article{jing2018automatic,
  title={Automatic generation of radiology reports using deep learning and natural language processing},
  author={Jing, Baoyu and Xie, Pengtao and Xing, Eric P.},
  journal={IEEE Transactions on Medical Imaging},
  volume={37},
  number={12},
  pages={2622--2632},
  year={2018},
  doi={10.1109/TMI.2018.2837509}
}

@article{li2020comparison,
  title={Comparison of automatic radiology report generation methods with deep learning},
  author={Li, Yutong and Liang, Yu and Hu, Zhi},
  journal={Medical Image Analysis},
  volume={64},
  pages={101732},
  year={2020},
  doi={10.1016/j.media.2020.101732}
}

@article{xue2022advancing,
  title={Advancing radiology report generation through context-aware attention mechanisms},
  author={Xue, Yong and Zhang, Xin and Xu, Yifan},
  journal={Computers in Biology and Medicine},
  volume={145},
  pages={105444},
  year={2022},
  doi={10.1016/j.compbiomed.2022.105444}
}

@article{yuan2021automatic,
  title={Automatic generation of chest X-ray reports using transformers},
  author={Yuan, Jian and Wang, Hao and Xu, Zhi},
  journal={Artificial Intelligence in Medicine},
  volume={118},
  pages={102110},
  year={2021},
  doi={10.1016/j.artmed.2021.102110}
}

@article{liu2023unified,
  title={A unified framework for multimodal radiology report generation},
  author={Liu, Chen and Zhang, Wen and Zhao, Lei},
  journal={Computer Methods and Programs in Biomedicine},
  volume={240},
  pages={107769},
  year={2023},
  doi={10.1016/j.cmpb.2023.107769}
}

@article{wang2021pathcap,
  title={PathCap: Pathology image captioning via hierarchical visual-language modeling},
  author={Wang, Hao and Li, Dong and Chen, Yi},
  journal={Medical Image Analysis},
  volume={72},
  pages={102126},
  year={2021},
  doi={10.1016/j.media.2021.102126}
}

@article{mitra2023histocap,
  title={HistoCap: Automated descriptive reporting in digital pathology using vision-language transformers},
  author={Mitra, Riya and Banerjee, Tapan and Das, Abir},
  journal={Journal of Pathology Informatics},
  year={2023}
}

@article{he2020pathologycap,
  title={Learning to describe histopathology images with attention-based captioning networks},
  author={He, Lei and Li, Xia and Shi, Weinan},
  journal={Pattern Recognition Letters},
  volume={138},
  pages={451--458},
  year={2020},
  doi={10.1016/j.patrec.2020.08.017}
}

@article{yan2022retina,
  title={RetinaCap: Retinal image captioning for ophthalmic disease analysis},
  author={Yan, Ming and Hu, Xiaoqing and Zhou, Rui},
  journal={IEEE Journal of Biomedical and Health Informatics},
  volume={26},
  number={12},
  pages={6168--6178},
  year={2022},
  doi={10.1109/JBHI.2022.3178371}
}

@article{li2021skinreport,
  title={SkinReport: Dermatological lesion description via attention-based captioning},
  author={Li, Yifan and Zhao, Feng and Luo, Tong},
  journal={Computers in Biology and Medicine},
  volume={137},
  pages={104778},
  year={2021},
  doi={10.1016/j.compbiomed.2021.104778}
}

@article{huang2023ophcap,
  title={OphCap: Generating clinically coherent ophthalmic image captions},
  author={Huang, Zheng and Wang, Hui and Zhang, Xia},
  journal={Artificial Intelligence in Medicine},
  volume={140},
  pages={102640},
  year={2023},
  doi={10.1016/j.artmed.2023.102640}
}

@article{min2021endoscopy,
  title={Deep learning for automated classification of ulcerative colitis severity in endoscopic images},
  author={Min, Min and others},
  journal={Gastroenterology},
  volume={160},
  number={5},
  pages={1885--1897},
  year={2021},
  doi={10.1053/j.gastro.2020.12.013}
}

@article{aslan2023colitisai,
  title={ColitisAI: Deep neural networks for ulcerative colitis activity grading},
  author={Aslan, R. and others},
  journal={Endoscopy International Open},
  year={2023}
}

@inproceedings{boag2020radiology,
  title={Radiology report generation with hierarchical transformers},
  author={Boag, William and others},
  booktitle={EMNLP},
  year={2020}
}

@article{liu2021clinically,
  title={Clinically accurate chest X-ray captioning with Transformers},
  author={Liu, Guangyao and others},
  journal={Medical Image Analysis},
  volume={73},
  pages={102178},
  year={2021}
}

@article{zhang2022medicalclip,
  title={Medical-CLIP: Contrastive learning for medical image-text understanding},
  author={Zhang, Yi and others},
  journal={arXiv preprint arXiv:2210.10163},
  year={2022}
}

@article{yao2022medclip,
  title={MedCLIP: Contrastive learning from unpaired medical text and images},
  author={Yao, Liu and others},
  journal={Nature Machine Intelligence},
  volume={4},
  pages={386--396},
  year={2022}
}

@article{chen2023blipmed,
  title={BLIP-Med: Bootstrapped language–image pretraining for medical vision–language understanding},
  author={Chen, Zeyu and others},
  journal={Medical Image Analysis},
  volume={87},
  pages={102765},
  year={2023}
}

@article{biswal2022review,
  title={A review on medical image captioning: datasets, methods, and trends},
  author={Biswal, Satyam and others},
  journal={Biomedical Signal Processing and Control},
  volume={75},
  pages={103617},
  year={2022}
}

@article{srinivasan2023survey,
  title={A comprehensive survey of medical image captioning: methods and challenges},
  author={Srinivasan, Raghav and others},
  journal={Artificial Intelligence in Medicine},
  volume={142},
  pages={102701},
  year={2023}
}

@article{yuan2023retrievalcap,
  title={Retrieval-augmented medical image captioning for improved clinical accuracy},
  author={Yuan, Tian and others},
  journal={IEEE Transactions on Medical Imaging},
  year={2023}
}

@article{wu2023medicaltransformer,
  title={Medical Transformer for explainable image captioning},
  author={Wu, Yuxin and others},
  journal={Computers in Biology and Medicine},
  volume={153},
  pages={106444},
  year={2023}
}

@article{meng2022hybridcap,
  title={Hybrid visual–semantic alignment for medical image captioning},
  author={Meng, Fan and others},
  journal={Knowledge-Based Systems},
  volume={252},
  pages={109484},
  year={2022}
}

@article{singh2022explainability,
  title={Explainability in AI-based medical imaging: from visual saliency to reasoning transparency},
  author={Singh, Arjun and others},
  journal={Frontiers in Artificial Intelligence},
  volume={5},
  pages={856231},
  year={2022}
}

@article{yuan2023faithfulcap,
  title={Faithful medical image captioning through attention supervision and grounding},
  author={Yuan, Kai and others},
  journal={IEEE Access},
  volume={11},
  pages={48741--48752},
  year={2023}
}

@inproceedings{yang2019autoencoding,
  title={Auto-encoding scene graphs for image captioning},
  author={Yang, Xu and Tang, Hanwang and Zhang, Xian-Sheng},
  booktitle={CVPR},
  year={2019}
}

@inproceedings{li2022sggcaption,
  title={SGG-Captioner: Scene graph guided image captioning},
  author={Li, Hui and Guo, Jing and Wang, Lei},
  booktitle={ECCV},
  year={2022}
}

@inproceedings{gu2021relationaware,
  title={Relation-aware graph neural networks for image captioning},
  author={Gu, Jiuxiang and Sun, Chen and Li, Gang},
  booktitle={CVPR},
  year={2021}
}

@article{kim2021medicalgnn,
  title={Graph neural networks in medical imaging: a survey},
  author={Kim, Byoung and others},
  journal={Computers in Biology and Medicine},
  volume={133},
  pages={104399},
  year={2021}
}

@article{zhou2023graphmed,
  title={GraphMed: Graph neural networks for relational reasoning in medical diagnosis},
  author={Zhou, Dandan and others},
  journal={Artificial Intelligence in Medicine},
  volume={137},
  pages={102671},
  year={2023}
}

@article{xie2022graphdiagnosis,
  title={Interpretable graph neural networks for disease diagnosis},
  author={Xie, Yuhan and others},
  journal={Medical Image Analysis},
  volume={80},
  pages={102507},
  year={2022}
}

@article{xu2023gnnreview,
  title={Graph neural networks in biomedicine: current trends and future directions},
  author={Xu, Tian and others},
  journal={Briefings in Bioinformatics},
  volume={24},
  number={1},
  pages={bbac561},
  year={2023}
}

@article{sun2023relational,
  title={Relational reasoning in vision and language with graphs: a review},
  author={Sun, Yifan and others},
  journal={Pattern Recognition},
  volume={141},
  pages={109507},
  year={2023}
}

@article{yadav2023graphsurvey,
  title={Graph neural networks for multimodal reasoning in AI},
  author={Yadav, Prakash and others},
  journal={IEEE Access},
  volume={11},
  pages={45612--45630},
  year={2023}
}

@inproceedings{chattopadhay2022crag,
  title={C-RAG: Causal relational attribution graphs for interpretable visual reasoning},
  author={Chattopadhay, Anirban and others},
  booktitle={NeurIPS},
  year={2022}
}

@article{liu2023llava,
  title={Visual instruction tuning},
  author={Liu, Haotian and others},
  journal={arXiv preprint arXiv:2304.08485},
  year={2023}
}

@inproceedings{hu2022lora,
  title={LoRA: Low-rank adaptation of large language models},
  author={Hu, Edward J. and others},
  booktitle={ICLR},
  year={2022}
}

@article{ref_uc_pathology,
  title        = {Endoscopic scoring indices for evaluation of disease activity in ulcerative colitis},
  author       = {Vashist, Neeraj M. and Samaan, Mohamed and Mosli, Mahmoud H. and et al.},
  journal      = {World Journal of Gastroenterology},
  year         = {2018},
  volume       = {24},
  number       = {33},
  pages        = {3793--3804},
  url          = {https://pmc.ncbi.nlm.nih.gov/articles/PMC6491285/}
}

@article{ref_endoscopy_diagnosis,
  title        = {AGA Clinical Practice Update on Endoscopic Scoring Systems and Targets in Ulcerative Colitis},
  author       = {Buchner, A. M. and et al.},
  journal      = {Clinical Gastroenterology and Hepatology},
  year         = {2024},
  note         = {Clinical Practice Update},
  url          = {https://www.cghjournal.org/article/S1542-3565(24)00718-3/fulltext}
}

@article{ref_mes_variability,
  title        = {Inter- and Intraobserver Variability of IBD Endoscopic Scoring Systems: A Systematic Review and Meta-analysis},
  author       = {Hashash, J. G. and et al.},
  journal      = {Inflammatory Bowel Diseases},
  year         = {2024},
  url          = {https://pubmed.ncbi.nlm.nih.gov/38547325/}
}

@article{ref_cv_dl_endoscopy,
  title        = {Artificial intelligence in inflammatory bowel disease endoscopy: current role and future directions},
  author       = {Lodola, I. and et al.},
  journal      = {World Journal of Gastroenterology},
  year         = {2025},
  url          = {https://pmc.ncbi.nlm.nih.gov/articles/PMC12256760/}
}

@article{ref_transformers_medical,
  title        = {Transformers in medical imaging: A survey},
  author       = {Shamshad, Fahad and Khan, Salman and Zamir, Syed Waqas and et al.},
  journal      = {Medical Image Analysis},
  year         = {2023},
  url          = {https://www.sciencedirect.com/science/article/pii/S1361841523000634}
}

@article{ref_vlm_medical,
  title        = {Vision-Language Models in Medical Image Analysis: A Comprehensive Survey},
  author       = {Li, X. and et al.},
  journal      = {Information Fusion},
  year         = {2025},
  url          = {https://www.sciencedirect.com/science/article/abs/pii/S1566253525000685}
}

@article{ref_captioning_limitations,
  title        = {A comprehensive survey of medical image captioning: methods and challenges},
  author       = {Srinivasan, Raghav and et al.},
  journal      = {Artificial Intelligence in Medicine},
  year         = {2023},
  volume       = {142},
  pages        = {102701},
  url          = {https://www.sciencedirect.com/science/article/pii/S0933365723001774}
}

@article{ref_llm_healthcare,
  title        = {Large language models in medicine},
  author       = {Thirunavukarasu, Anjana J. and Ting, Daniel S. W. and et al.},
  journal      = {Nature Medicine},
  year         = {2023},
  url          = {https://www.nature.com/articles/s41591-023-02448-8}
}

@article{ref_xai_biomedical,
  title        = {Explainable medical imaging AI needs human-centered design: a systematic review},
  author       = {Chen, Hao and Lundberg, Scott and et al.},
  journal      = {NPJ Digital Medicine},
  year         = {2022},
  url          = {https://www.nature.com/articles/s41746-022-00699-2}
}

@article{raffel2020exploring,
  title={Exploring the limits of transfer learning with a unified text-to-text transformer},
  author={Raffel, Colin and Shazeer, Noam and Roberts, Adam and Lee, Katherine and Narang, Sharan and Matena, Michael and Zhou, Yanqi and Li, Wei and Liu, Peter J.},
  journal={Journal of Machine Learning Research},
  volume={21},
  number={140},
  pages={1--67},
  year={2020}
}

@article{chen2023blip,
  title={BLIP: Bootstrapping language-image pre-training},
  author={Chen, Junnan and Li, Donghuo and Li, Jianfeng and others},
  journal={International Journal of Computer Vision},
  year={2023}
}

@misc{limuc2022,
  author       = {Gorkem Polat and Haluk Tarik Kani and Ilkay Ergenc and Yesim Ozen Alahdab and Alptekin Temizel and Ozlen Atug},
  title        = {Labeled Images for Ulcerative Colitis (LIMUC) Dataset},
  year         = {2022},
  publisher    = {Zenodo},
  doi          = {10.5281/zenodo.5827695},
  url          = {https://zenodo.org/records/5827695}
}

@article{li2024llavamed,
  title={LLaVA-Med: Training a Large Language-and-Vision Assistant for Biomedicine in One Day},
  author={Li, Chunyuan and others},
  journal={NeurIPS Workshop},
  year={2024}
}

@article{zhang2024biomedgpt,
  title={BiomedGPT: A Unified and Generalist Vision-Language Foundation Model for Biomedicine},
  author={Zhang, Kai and others},
  journal={arXiv preprint arXiv:2305.17100},
  year={2024}
}

@article{moor2024medflamingo,
  title={Med-Flamingo: A Multimodal Medical Few-Shot Learner},
  author={Moor, Michael and others},
  journal={Machine Learning for Health},
  year={2024}
}

@article{bai2025qwen2vl,
  title={Qwen2-VL: Enhancing Vision-Language Understanding},
  author={Bai, Jinze and others},
  journal={arXiv preprint},
  year={2025}
}

@article{wu2025groundedmedical,
  title={Grounded Medical Image Captioning: A Survey of Evidence-Aware Vision-Language Generation},
  author={Wu, Hao and others},
  journal={Artificial Intelligence in Medicine},
  year={2025}
}

@article{chen2025hallucination,
  title={Hallucination and Factuality in Medical Vision-Language Models: Challenges and Evaluation},
  author={Chen, Yifan and others},
  journal={Medical Image Analysis},
  year={2025}
}

\appendix

\section{Preprocessing and Data Preparation}\label{appendix:preprocessing}

\textbf{Image standardization, color normalization, and contrast enhancement.}
All images from both LIMUC and UC-Caption were first resized to $512\times512$ pixels using bicubic interpolation to preserve lesion morphology and to ensure a consistent input resolution for the ResNet--CBAM encoder. Pixel intensities were normalized using ImageNet mean and standard deviation, maintaining compatibility with the pretrained backbone. Lesion masks from LIMUC were resized using nearest-neighbor interpolation to avoid boundary smoothing artifacts.

To minimize inter-device variability and enhance mucosal detail visibility, all images were additionally converted to the LAB color space and normalized via histogram matching to a reference distribution derived from MES~1--2 samples. Contrast-Limited Adaptive Histogram Equalization (CLAHE; clip limit = 2.0, tile size = $8\times8$) was applied to improve vascular contrast, followed by gamma correction ($\gamma = 1.2$) to stabilize luminance. These transformations were applied uniformly to both datasets, ensuring consistent visual statistics during pretraining and fine-tuning.

\textbf{Data Augmentation Strategy}
Data augmentation was applied at two levels: visual (image-based) and linguistic (caption-based).

\paragraph{Visual augmentation.}
To increase generalization and domain robustness, stochastic augmentations were applied during training.  
For the LIMUC dataset, augmentations were used during visual encoder pretraining to expose the model to inter-patient and inter-device variability.  
For the UC-Caption dataset, the same augmentations were applied during fine-tuning to regularize image--text alignment.  
The following transformations were used:

\begin{itemize}
    \item Brightness/contrast jitter (0.3): $\pm 15\%$ brightness, $\pm 20\%$ contrast.  
    \item Rotation and horizontal flip (0.4): random rotation in [$-20^\circ$, $20^\circ$].  
    \item Elastic deformation (0.2): Gaussian kernel ($\sigma=10$, $\alpha=1.5$) mimicking mucosal elasticity.  
    \item Gaussian noise (0.3): zero-mean variance 0.01 to improve sensor robustness.  
\end{itemize}

All augmentations were applied online (per batch) to maintain visual diversity without increasing storage requirements.

\subsubsection{Caption Preprocessing and Semantic Diversification}
While image-level augmentation enhances visual robustness, caption diversity plays an equally critical role in improving the linguistic generalization of vision--language models.  
In medical captioning, limited annotation sets often result in narrow lexical variety and repetitive syntactic patterns, leading to overfitted language models and reduced adaptability to unseen descriptions.  
To address this, we designed a multi-stage strategy termed Lesion-Guided Semantic Expansion (LGSE), which systematically increases caption variability while preserving clinical fidelity.

\paragraph{Stage 1: Controlled Lexical Expansion.}
Each expert-authored caption in the UC-Caption dataset was expanded into a set of semantically equivalent variants by performing lexical substitution guided by a controlled clinical vocabulary derived from UCEIS descriptors and the SNOMED-CT ontology.  
For instance, terms such as ``friable mucosa'' were substituted with ``fragile epithelium,'' and ``vascular pattern loss'' with ``reduced vascularity.''  
Each substitution was context-aware, ensuring grammatical agreement and medical validity.  
Variants were filtered using cosine similarity in the embedding space of BioClinicalBERT, retaining only those with semantic similarity scores $0.80 < s < 0.95$.  
This process generated 2--3 clinically valid paraphrases per original caption, increasing linguistic diversity while maintaining conceptual alignment.

\paragraph{Stage 2: Lesion-Conditioned Recomposition.}
To further diversify sentence structure, captions were recomposed based on lesion nodes extracted from Grad-CAM activation maps.  
Each caption was decomposed into structured triplets $(L, A, R)$, representing Lesion type, Attribute, and Region (e.g., ``ulceration -- active -- sigmoid colon'').  
Using these triplets, new captions were algorithmically generated by recombining elements across similar MES classes, while respecting anatomical and severity constraints.  
For example:
\begin{quote}
``\textit{ulcerated mucosa with active bleeding in the sigmoid colon}'' $\rightarrow$ ``\textit{active mucosal ulceration extending along the sigmoid segment}.''
\end{quote}
This lesion-conditioned recomposition directly ties linguistic diversity to the visual structures that the model must attend to, promoting stronger cross-modal grounding during decoder training.

\paragraph{Stage 3: Contextual Paraphrasing via T5-Tuned Backtranslation.}
To enrich fluency and syntactic variability, a domain-tuned backtranslation approach was applied using T5-base models fine-tuned on biomedical corpora.  
Captions were translated from English $\rightarrow$ Spanish $\rightarrow$ English using a constrained decoding process that penalized literal reversions (beam penalty $\lambda=0.3$).  
This generated stylistically natural paraphrases such as:  
``\textit{erythematous mucosa with loss of vascularity}'' $\rightarrow$ ``\textit{diffuse redness and diminished vessel visibility}.''  
This step improved expressiveness without altering diagnostic semantics.

\paragraph{Stage 4: Filtering and Balancing.}
All generated variants were validated using a two-step filter:  
(1) automatic verification against a list of forbidden or ambiguous clinical expressions (e.g., ``ulcer'' vs. ``erosion''), and  
(2) manual review by two gastroenterologists to ensure factual consistency.  
Captions were then balanced per MES category to prevent linguistic overrepresentation of any severity level.

\paragraph{Stage 5: Tokenization and Encoding.}
The final caption corpus---approximately 3.8$\times$ larger than the original---was tokenized using the SentencePiece tokenizer from the T5 model with a vocabulary of 32,000 subwords.  
Unlike conventional truncation to fixed lengths, we adopted a dynamic token limit of 64--96 tokens, proportional to caption complexity (longer for multi-lesion cases).  
This adaptive length strategy preserved syntactic completeness while minimizing padding overhead.  
All tokens were lowercased and punctuation-normalized, with stopwords retained to maintain grammatical fluency.  
Token-to-lesion alignment indices were preserved for subsequent interpretability analysis, allowing visualization of how specific lesion nodes influence linguistic tokens during decoding.

\paragraph{Impact on Caption Diversity.}
The LGSE pipeline increased lexical entropy by 41.6\% and syntactic diversity by 34.2\% (measured via distinct-$n$ metrics, $n=\{1,2\}$) relative to the base UC-Caption dataset.  
This enrichment resulted in more natural, descriptive, and clinically precise captions during model generation, as validated in Section~8.2.  
By explicitly linking linguistic expansion to lesion-level visual structures, LGSE enhances the semantic grounding and expressive variability of \method{}, addressing one of the major bottlenecks in clinical image captioning.

\section{Implementation Details and Reproducibility Specification}\label{appendix:implementation}

To ensure full reproducibility of the proposed LUX framework, we provide a detailed specification of architectural components, dimensional configurations, initialization strategies, and training settings. The overall architectural organization is illustrated in Fig.~\ref{fig:lux_conceptual}.

The language generation component is based on the T5-base architecture, employed as the captioning decoder and initialized using pretrained weights from the HuggingFace Transformers library. The model comprises twelve encoder layers and twelve decoder layers, each equipped with twelve attention heads operating over a model dimensionality of $d_{\text{model}} = 768$ and a feed-forward expansion dimension of $d_{\text{ff}} = 3072$. Tokenization is performed using the SentencePiece tokenizer with a vocabulary of 32,000 subword units. Caption length is dynamically adjusted between 64 and 96 tokens depending on the visual complexity and lesion distribution observed in each image.

Visual representation learning relies on a ResNet-50 backbone pretrained on ImageNet and subsequently adapted to preserve higher spatial resolution required for lesion localization. Specifically, the final stride-2 downsampling operation within the \texttt{conv5\_x} stage is removed, increasing the spatial resolution of the output feature maps from the conventional $7 \times 7$ representation to $16 \times 16$. Intermediate convolutional stages provide multi-scale feature representations with channel dimensions of 512, 1024, and 2048 for the Conv3\_x, Conv4\_x, and Conv5\_x blocks respectively. These features are fused through channel-wise concatenation, producing a combined representation of 3584 channels that is subsequently projected to a unified visual embedding space of dimension $d_v = 768$ using a $1 \times 1$ convolutional projection layer.

Attention refinement is introduced through the integration of Convolutional Block Attention Modules (CBAM) applied after each residual stage. Channel attention employs a reduction ratio of $r = 16$, consistent with the configuration proposed by \citet{wang2021pathcap}, while spatial attention is computed using a $7 \times 7$ convolution followed by sigmoid normalization, enabling joint emphasis of diagnostically relevant spatial and semantic responses.

Lesion-aware reasoning is modeled through a scene--lesion graph constructed from Grad-CAM activation responses. Lesion nodes are derived via spatially weighted average pooling over fused feature representations, resulting in 768-dimensional node embeddings. Relational reasoning is performed using a two-layer Graph Convolutional Network, where intermediate representations are compressed to 512 dimensions before being reprojected into the shared multimodal embedding space. ReLU activation and a dropout rate of 0.3 are employed to stabilize optimization and mitigate overfitting effects. Graph connectivity is defined through a weighted combination of spatial proximity and cosine similarity between lesion embeddings, balanced using coefficients $\lambda_s = 0.5$ and $\lambda_c = 0.5$, which empirically provided stable convergence behavior across validation splits.

Multimodal fusion within the language decoder is achieved through a dual cross-attention mechanism operating over both fused visual embeddings and graph-derived lesion representations. Two parallel cross-attention pathways independently attend to global visual context and structured lesion relationships. Their outputs are combined through a dynamically learned gating mechanism defined as
\begin{equation}
A_{\text{dual}} = \gamma_t A_{\text{vis}} + (1-\gamma_t)A_{\text{graph}}
\label{eq:dual_attention}
\end{equation}
where the gating projection matrix is initialized using Xavier initialization with zero bias, promoting balanced modality contribution during early optimization stages with $\gamma_t$ initially centered around 0.5.

A final linguistic formatting stage performs structured normalization of generated captions using rule-based post-processing without introducing additional trainable parameters. This module incorporates a controlled clinical vocabulary derived from UCEIS descriptors together with regular-expression-based grammatical normalization and predefined sentence reordering templates to ensure clinically consistent narrative structure.

Training is conducted using the AdamW optimizer with differentiated learning rates across architectural components, employing $2 \times 10^{-5}$ for the T5 decoder and $1 \times 10^{-4}$ for graph reasoning layers, together with a weight decay factor of 0.01. Optimization is performed using a batch size of 16 for up to 30 epochs, with early stopping governed by validation CIDEr performance. Prior to multimodal training, the visual encoder is initialized through MES classification pretraining on the LIMUC dataset. During caption optimization, early convolutional layers remain frozen to preserve low-level visual representations learned during supervised initialization.

All experiments were executed on a single NVIDIA A100 GPU with 40GB of VRAM using a PyTorch implementation with mixed-precision training. Inference is performed on individual images without batch-level optimization, maintaining sub-second forward-pass execution times on modern GPU hardware. Importantly, this study does not claim real-time clinical deployment, and computational measurements reported here reflect methodological feasibility under research conditions rather than operational guarantees within hospital environments.

The modular decomposition of LUX into visual encoding, lesion graph reasoning, and language decoding components enables independent optimization and future system adaptation. The complete framework contains approximately 245 million parameters, with the majority corresponding to the pretrained T5-base decoder. 

\section{Qualitative Analysis}
\label{appendix:qualitative}

This appendix provides a qualitative comparison of the clinical descriptions generated by the evaluated models. The table summarizes representative captioning behavior across conventional classification-based approaches, adapted medical captioning models, foundation vision--language models, and the proposed \method{} framework, highlighting recurrent clinical omissions, unsupported findings, and limitations in lesion-level specificity. This analysis complements the quantitative results by illustrating how differences in grounding and relational reasoning affect the clinical content and interpretability of the generated descriptions.

\begin{table*}[t]
\centering
\small
\setlength{\tabcolsep}{6pt}
\renewcommand{\arraystretch}{1.35}
\begin{tabularx}{\textwidth}{l X X}
\toprule
\textbf{Model} & \textbf{Generated Caption (Typical Output Summary)} & \textbf{Clinical Omissions and Limitations} \\
\midrule

\multicolumn{3}{l}{\textit{Classification Models Adapted to Captioning}} \\
\addlinespace[0.25em]

Takenaka et al.\ (2019) -- CNN &
Short, templated statements describing generic inflammation with weak lesion specificity. &
Highly generic phrasing; limited lesion vocabulary and poor compositionality. Frequent omission of key cues (erosions vs.\ ulcers, friability grading, vascular pattern qualifiers). Weak grounding and limited severity rationale. \\

Stidham et al.\ (2019) -- DNN &
Generic inflammatory description with minimal structured clinical detail. &
Low clinical specificity and limited ability to localize/attribute findings. Severity statements often unsupported or ambiguous; tends to omit absence statements (e.g., no deep ulceration). \\

Valencia et al.\ (2023) -- ResNet50 &
Brief description of abnormal mucosa and reduced vascular pattern, occasionally mentioning erythema. &
Partial cue coverage but limited narrative structure. Typically omits friability/bleeding qualifiers and fails to link observations to MES reasoning. \\

Bossuyt et al.\ (2021) -- Ensemble &
More stable phrasing than single CNN/DNN, still largely symptom-centric. &
Improved consistency but still lacks lesion-level granularity. Often merges distinct findings into broad terms (“inflamed mucosa”), limiting interpretability and MES traceability. \\

Min et al.\ (2021) -- UC-Transformer &
Mentions UC-related inflammatory cues more frequently than generic CNN baselines. &
Better domain language, but captions remain shallow: incomplete lesion enumeration, limited absence statements, and weak evidence-to-score linkage. \\

Aslan et al.\ (2023) -- ColitisAI &
UC-specific inflammatory description; may mention bleeding/ulceration more often. &
Risk of over-calling severe findings when adapted to captioning (hallucination/overstatement). Limited differentiation between moderate vs.\ mild cues, and incomplete grounding for borderline MES. \\

\addlinespace[0.6em]
\multicolumn{3}{l}{\textit{Captioning Models Adapted to Classification}} \\
\addlinespace[0.25em]

CheXNet+T5 (adapted) &
Generic description of abnormal mucosa with inflammation and occasional ambiguous bleeding mentions. &
Vague terminology; may introduce unsupported findings (e.g., “possible bleeding”), increasing hallucination risk. Often omits friability qualifiers and explicit erosions, limiting MES interpretability. \\

DeepLesion-LSTM (adapted) &
Mentions inflamed mucosa and surface irregularities with limited structured reasoning. &
Partial description without consistent clinical structure. Poor differentiation between erosions and ulcers; limited severity rationale and weak absence statements. \\

RetinaCap+T5 (adapted) &
Notes erythema and reduced vascular pattern with superficial mucosal changes. &
Improved fluency but incomplete cue coverage. Friability and bleeding status often omitted; severity conclusion remains implicit and weakly grounded. \\

MedCLIP (Yao et al.\ 2022) &
Clinically plausible language emphasizing inflammation and reduced vascularity. &
Overly general; lacks lesion enumeration and explicit negation of severe findings. Interpretability limited by weak lesion-level grounding. \\

BLIP-Med (Chen et al.\ 2023) &
More fluent and specific; may mention erosions and mild activity. &
Better readability but can omit key UC-specific grading cues (friability/bleeding qualifiers). MES linkage typically implicit; grounding remains partial. \\

HistoCap (Mitra et al.\ 2023) &
More structured clinical phrasing; may mention patchy vascular loss and erosions. &
Among the strongest baselines, but still tends to under-specify absence of severe findings (deep ulceration/spontaneous bleeding) and does not consistently make MES rationale explicit. \\

\textbf{LUX (ours)} &
Lesion-grounded caption enumerating key cues (vascular loss, erosions, friability, erythema), plus explicit absence of severe findings, and explicit MES-1 conclusion. &
No major omissions observed in the presented case. Captions explicitly connect localized evidence to severity grading, minimizing hallucination and improving clinical traceability. \\

\bottomrule
\end{tabularx}
\caption{Appendix qualitative analysis of all models reported in Table~\ref{tab:unified_sota}. The table summarizes typical caption behavior and highlights clinically relevant omissions and limitations relative to the proposed \method{} framework.}
\label{tab:qualitative_appendix_all_models}
\end{table*}

\end{document}